%% file: main.tex
\documentclass{article}

 \usepackage[main, final, nonatbib]{neurips_2025}

\usepackage[utf8]{inputenc} % allow utf-8 input
\usepackage[T1]{fontenc}    % use 8-bit T1 fonts
\usepackage{hyperref}       % hyperlinks
\usepackage{url}            % simple URL typesetting
\usepackage{booktabs}       % professional-quality tables
\usepackage{amsfonts}       % blackboard math symbols
\usepackage{nicefrac}       % compact symbols for 1/2, etc.
\usepackage{microtype}      % microtypography
\usepackage{xcolor}         % colors

\usepackage{amssymb}
\usepackage{mathtools}
\usepackage{graphicx} 
\usepackage{amsmath}
\usepackage{amsthm}
\usepackage{amsfonts}
\usepackage{xspace} 
\usepackage{algorithm}
\usepackage{algorithmic}
\usepackage{color}
\usepackage{multirow}
\usepackage{booktabs}

\usepackage{multibib}
\newcites{newcite}{Reference}

\newcommand{\ouralg}{\texttt{FairOBD}\xspace}
\newcommand{\opt}{\texttt{OPT}\xspace}
\newcommand{\fairopt}{\texttt{FairOPT}\xspace}
\newcommand{\hitmin}{\texttt{HITMIN}\xspace}
\newcommand{\robd}{\texttt{ROBD}\xspace}
\newcommand{\dmd}{\texttt{DMD}\xspace}

\theoremstyle{plain}
\newtheorem{theorem}{Theorem}[section]

\newtheorem{lemma}[theorem]{Lemma}

\theoremstyle{definition}
\newtheorem{definition}[theorem]{Definition}
\newtheorem{assumption}[theorem]{Assumption}
\theoremstyle{remark}

\title{Fairness-Regularized Online Optimization with Switching Costs}

\author{
  Pengfei Li \\
  School of Information\\
  Rochester Institute of Technology\\
  \texttt{pflics@rit.edu} \\
  \And
  Yuelin Han \\
  Electrical and Computer Engineering\\
  University of California, Riverside \\
  \texttt{yhan116@ucr.edu}
  \And
  Adam Wierman \\
  Computing \& Mathematical Sciences\\ California Institute of Technology \\
  \texttt{adamw@caltech.edu} \\
  \And
  Shaolei Ren \\
  Electrical and Computer Engineering\\
  University of California, Riverside \\
  \texttt{shaolei@ucr.edu} 
}

\begin{document}

\maketitle

\begin{abstract}
Fairness and action smoothness are two crucial considerations in many online optimization problems, but they have yet to be addressed simultaneously.
In this paper, we study a new and challenging setting of fairness-regularized smoothed online convex optimization
with switching costs.
First, to highlight
the fundamental challenges introduced by the long-term fairness
regularizer evaluated based on the entire sequence
of actions, we prove that even without switching costs,
no online algorithms can possibly achieve a sublinear regret
or finite competitive ratio compared to the offline optimal algorithm as
the problem episode length $T$ increases.
Then, we propose \ouralg (Fairness-regularized
Online Balanced Descent), which reconciles the tension between minimizing the hitting
cost, switching cost, and fairness cost.
Concretely, \ouralg decomposes the long-term fairness cost into a sequence
of online costs by introducing
an auxiliary variable and then leverages
the auxiliary variable
to regularize the online actions for fair outcomes. Based on a new approach to account for switching costs, we prove that \ouralg offers a worst-case asymptotic competitive
ratio against a novel benchmark---the optimal offline algorithm with parameterized constraints---
by considering  $T\to\infty$. Finally, we run trace-driven experiments of dynamic computing resource provisioning for socially responsible AI inference to empirically evaluate  \ouralg, showing that \ouralg can effectively reduce
the total fairness-regularized cost and better promote fair outcomes compared
to existing baseline solutions.
\end{abstract}

\input{tex/intro}

\input{tex/related}

\input{tex/problem_definition}

\input{tex/hardness_long_term}

\input{tex/performance}
\input{tex/algorithm}

\input{tex/simulation}
\input{tex/conclusion}

\bibliographystyle{plain}
\bibliography{main.bbl}

\newpage
\appendix

\section*{Appendix}
\input{tex/appendix_application}

\input{tex/appendix_simulation}

\input{tex/appendix_proof_lb}

\input{tex/appendix_proof_acr}

\clearpage
\bibliographystylenewcite{plain}
\bibliographynewcite{newcite.bbl}

\end{document}

%% file: tex/intro.tex
\section{Introduction}

Smoothed decision-making is critical to reducing abrupt and even potentially dangerous large action changes in many online optimization problems, e.g., energy production scheduling in power grids, object tracking, motion planning, and server capacity provisioning
in data centers, among others \cite{SOCO_Memory_Adam_NIPS_2020_NEURIPS2020_ed46558a,SOCO_Revisiting_Nanjing_NIPS_2021_zhang2021revisiting,Bandit_Switching_NeurIPS_2022_NEURIPS2022_6590cb82}. Mathematically, 
action smoothness can be effectively achieved by including into the optimization objective a switching cost that penalizes temporal changes in actions.
The added switching cost essentially equips the online optimization objective with a (finite) memory
of the previous actions and also creates substantial algorithmic challenges.
As such, it has received a huge amount of attention in the last
decade, with a quickly growing list of algorithms developed under
various settings 
 (see \cite{Bandit_Switching_Constrained_BoJi_VT_INFOCOM_2023_10228986,SOCO_MetricUntrustedPrediction_Google_ICML_2020_pmlr-v119-antoniadis20a,SOCO_DynamicRightSizing_Adam_Infocom_2011_LinWiermanAndrewThereska,SOCO_CapacityScalingAdaptiveBalancedGradient_Gatech_MAMA_2021_Sigmetrics_2021_10.1145/3512798.3512808,SOCO_Prediction_LinearSystem_NaLi_Harvard_NIPS_2019_10.5555/3454287.3455620,SOCO_Revisiting_Nanjing_NIPS_2021_zhang2021revisiting,SOCO_Memory_Adam_NIPS_2020_NEURIPS2020_ed46558a,SOCO_ML_ChasingConvexBodiesFunction_Adam_COLT_2022,SOCO_NonConvex_Adam_Sigmetrics_2020_10.1145/3379484,SOCO_OBD_Niangjun_Adam_COLT_2018_DBLP:conf/colt/ChenGW18} and the references therein).

Additionally, (long-term) fairness is a crucial consideration that must be carefully addressed in a variety
of online optimization problems, especially those
that can profoundly impact individuals' well-being and social welfare. For example, sequential allocation of social goods
to different individuals/groups must be fair without discrimination \cite{Fairness_OnlineAllocation_Regularized_Fairness_DMD_arXiv_2020_balseiro2021regularized,Fair_Sequential_OptimalEnvyEfficieny_Sid_ChristinaYu_Cornell_OR_2023_10.1287/opre.2022.2397}, 
the surging environmental costs
of geographically distributed AI data centers must be fairly
distributed across different regions as mirrored
in recent repeated calls by various organizations and government agencies 
\cite{Justice_AINowInstitute_ConfrontingTechPower_2023,Justice_AI_EnvironmentalEquity_CaliforniaReport_2023,Shaolei_Equity_GLB_Environmental_AI_eEnergy_2024,Justice_WaterMarginalized_BrookingsInstitution_2024},
and AI model performance and resource provisioning must promote social fairness without adversely impacting certain disadvantaged individuals/groups \cite{Fair_ML_Survey_ACM_2022_10.1145/3494672}.
If not designed with fairness in mind, online algorithms can further perpetuate or even exacerbate societal biases, which can disproportionately affect certain groups/individuals and amplify the already widening socioeconomic disparities. 
Therefore,
fairness in online optimization is not just a matter of ethical concern but also a necessity for ensuring broader social responsibility.

The incorporation of a long-term fairness regularizer into the objective function can effectively
achieve fairness in online optimization
by
mitigating implicit or explicit biases 
that could otherwise be reinforced \cite{Fairness_OnlineAllocation_Regularized_Fairness_DMD_arXiv_2020_balseiro2021regularized,Neely_Booklet,Neely_Universal,Shaolei_Equity_GLB_Environmental_AI_eEnergy_2024}. Despite its effectiveness in promoting fairness, the added fairness regularizer presents significant challenges for online optimization\cite{Fairness_OnlineAllocation_Regularized_Fairness_DMD_arXiv_2020_balseiro2021regularized,Neely_Booklet}. This arises from the lack of complete future information, which is required to evaluate the fairness regularizer. Intuitively, online decision-makers prioritize immediate cost reduction, reacting to sequentially revealed information, potentially neglecting the long-term consequences of the decision sequence.

Recent research has started to explore fairness in various online problems, such as online resource allocation, using primal-dual techniques \cite{Neely_Universal,Fairness_OnlineAllocation_Regularized_Fairness_DMD_arXiv_2020_balseiro2021regularized, HKUST_DMD_resolving_ma2025optimal}. However, a major limitation of current fair online algorithms lies in their reliance on the assumption that per-round objective functions are   
independent of historical actions.
This assumption plays a pivotal role in both algorithm design and theoretical analysis (e.g., \cite{Fairness_OnlineAllocation_Regularized_Fairness_DMD_arXiv_2020_balseiro2021regularized, OnlineAllocation_DualMirroDescent_Google_OperationalResearch_2022_doi:10.1287/opre.2021.2242}). 
For instance, in a simple resource allocation scenario, while the maximum allowable action at time $t$ and its corresponding reward may depend on past allocation decisions through the remaining budget,  
the revenue or reward function at time $t$ is assumed to depend only
on the current allocation decision and remain independent of these historical decisions.
However, these methods fall short in addressing online problems where the per-round objective function is directly impacted by historical actions, such as those involving switching costs, as detailed in Section~\ref{sec:related}.
 
\textbf{Contributions.} In this paper, we study fairness-regularized smoothed online optimization
with switching costs, a new and challenging setting that has not been considered in the literature to our knowledge. More specifically, we incorporate a fairness regularizer that captures the long-term impacts of online actions,
and aim to minimize the sum of a per-round hitting cost, switching cost,
and a long-term fairness cost.  
To simultaneously address the tension between minimizing the hitting
cost, switching cost, and fairness cost, we propose \ouralg (Fairness-regularized
Online Balanced Descent).

Optimizing the long-term cost with a sequentially revealed context sequence is widely recognized as a challenging problem (e.g., \cite{Neely_Universal, Fairness_OnlineAllocation_Regularized_Fairness_DMD_arXiv_2020_balseiro2021regularized, OnlineAllocation_DualMirroDescent_Google_OperationalResearch_2022_doi:10.1287/opre.2021.2242, network_fair_dmd_bandit_feedback_chen_2024,   fairness_unobserved_context_inria_neurips_2023, learning_schedule_competitive_fairness_eenergy_pengfei}). A common strategy to address this challenge involves decomposing the long-term fairness cost into a sequence of online costs by introducing an auxiliary variable (e.g. \cite{dual_mirror_descent_balseiro2023best, neely_Lyapunov_boook_10.5555/1941130, HKUST_DMD_resolving_ma2025optimal}, which effectively converts the fairness cost into an equivalent long-term constraint.
However, the key limitation of existing methods is the requirement of  
independence between per-round objective functions, which is violated by the switching cost. 
Specifically, given the updated Lagrangian multiplier, the per-round online objective  depends on the previous \textit{irrevocable} action due to the switching cost.
This dependency is not permitted in prior fairness-regularized online problems \cite{Fairness_OnlineAllocation_Regularized_Fairness_DMD_arXiv_2020_balseiro2021regularized,Neely_Universal}.
As a result, due to the differing action sequences, both \ouralg and the benchmark algorithm $(R, \delta)$-OPT (Definition~\ref{def:opt_R_delta}) have distinct objective functions at each round, complicating a direct comparison of their costs.

In this work, we rigorously analyze the inherent hardness of incorporating long-term fairness regularization into online optimization. 
We establish fundamental lower bounds in the dynamic benchmark under two standard performance metrics: regret and competitive ratio.
Unlike the static benchmark, where the offline optimal action is constrained to a single fixed action, the dynamic benchmark allows the offline optimal action sequence to adapt over time with full knowledge of the context. 
While a finite competitive ratio is attainable in online optimization with switching costs, whether a similar finite competitive ratio can be achieved in presence of long-term fairness regularization remains an open question. 
Intuitively, this is because long-term regularizer depend on the entire trajectory and can only be evaluated at the end of an episode, whereas switching costs can be assessed step by step. 
Although prior work has made conjectures about the impossibility of achieving finite performance in similar settings \cite{Fairness_OnlineAllocation_Regularized_Fairness_DMD_arXiv_2020_balseiro2021regularized}, our results provide the first rigorous proof. Our results show that in the adversarial setting, all online algorithms inevitably incur a constant regret gap (leading to linear total regret) and an unbounded competitive ratio that grows at least linearly with the horizon length. 
This demonstrates that the unconstrained dynamic benchmark is unsuitable for analyzing such problems. 
Motivated by real-world applications and guided by insights from our proofs, we propose the $(R, \delta)$-benchmark, in which the deviation between piece-wise and episode-wise function averages is bounded.

By developing a novel proof technique that explicitly accounts for the dependency
of online objectives on previous actions,
we show that \ouralg offers a worst-case asymptotic competitive
ratio against a $(R,\delta)$-constrained optimal offline algorithm (Theorem~\ref{thm:dmd_algorithm_switch} in Section~\ref{sec:analysis}), where $R$ and $\delta$ specify the total allowed variations of the optimal actions. 
Importantly, our analysis quantifies the impact of the long-term fairness regularizer on the convergence speed of \ouralg's total cost, which asymptotically matches the best-known results for two existing settings -- smoothed online optimization without fairness and fairness-regularized online optimization without switching costs—as special cases. This demonstrates the power and generality of \ouralg. 

Finally, to empirically evaluate the performance of \ouralg, 
we conduct trace-driven experiments for AI workload shifting with a focus on geographically balanced distribution of public health risks.
Our results demonstrate that \ouralg can effectively reduce
the total fairness-regularized cost, highlighting the empirical advantages of \ouralg compared
to existing baseline solutions.

In summary, our impossibility results and the proposed $(R, \delta)$ benchmark lay a solid foundation for studying online optimization with long-term fairness. We further advance the literature on smoothed online optimization by introducing a novel fairness regularizer that promotes equitable outcomes.
Most importantly, we overcome the key technical challenges of entangling switching costs as a type of action memory into the mirror descent update process, which is not allowed by the prior fairness-regularized online algorithms \cite{Fairness_OnlineAllocation_Regularized_Fairness_DMD_arXiv_2020_balseiro2021regularized,Neely_Universal} or
smoothed online algorithms \cite{SOCO_ROBD_Adam_NeurIPS_2019_10.5555/3454287.3454455}.

%% file: tex/related.tex
\section{Related Works}\label{sec:related}

\textbf{Smoothed online optimization.}
Smoothed online optimization is a challenging problem for which a growing list
of algorithms have been designed to offer the worst-case performance guarantees
in both adversarial and stochastic settings
\cite{SOCO_ML_ChasingConvexBodiesFunction_Adam_COLT_2022,SOCO_Memory_Adam_NIPS_2020_NEURIPS2020_ed46558a,SOCO_OBD_LQR_Abstract_Goel_Adam_Caltech_2019_10.1145/3374888.3374892,SOCO_Revisiting_Nanjing_NIPS_2021_zhang2021revisiting,SOCO_OBD_Niangjun_Adam_COLT_2018_DBLP:conf/colt/ChenGW18}. 
Recently, it has also been extended in various directions, including
decentralized networked systems \cite{Decentralized_SOCO_YihengLin_Adam_ICML_2022_pmlr-v162-lin22c},
hitting cost feedback delays  \cite{SOCO_Memory_FeedbackDelay_Nonlinear_Adam_Sigmetrics_2022_10.1145/3508037},
bandit cost feedback \cite{Bandit_Switching_NeurIPS_2022_NEURIPS2022_6590cb82,Bandit_Switching_Constrained_BoJi_VT_INFOCOM_2023_10228986,Bandit_Switching_Regret_STOC_2014_10.1145/2591796.2591868}, switching cost constraints \cite{SOCO_OCO_SwitchingBudget_LazyOCO_COLT_2021_pmlr-v134-sherman21a,SOCO_OCO_ContinuousSwitchingConstraints_LijunZhang_Nanjing_NeurIPS_2021_wang2021online},
future cost predictions \cite{SOCO_Prediction_Error_RHIG_NaLi_Harvard_NIPS_2020_NEURIPS2020_a6e4f250,SOCO_Prediction_LinearSystem_NaLi_Harvard_NIPS_2019_10.5555/3454287.3455620},
and machine learning advice augmentation \cite{SOCO_ML_ChasingConvexBodiesFunction_Adam_COLT_2022,SOCO_MetricUntrustedPrediction_Google_ICML_2020_pmlr-v119-antoniadis20a,SOCO_OnlineOpt_UnreliablePrediction_Adam_Sigmetrics_2023_10.1145/3579442,Shaolei_Learning_SOCO_Delay_NeurIPS_2023}, among others.

By considering a dynamic (constrained) offline
optimal algorithm as the benchmark, our work focuses on the worst-case competitive ratio and considers the classic smoothed online convex optimization setting   \cite{SOCO_ROBD_Adam_NeurIPS_2019_10.5555/3454287.3454455},
but adds a long-term horizon fairness regularizer. This addition
presents significant challenges to our algorithm design and is the
key novelty that separates our work from the rich literature on smoothed 
online optimization. Another study
\cite{SOCO_SimpleLongTermConstraint_UMass_AdamWierman_Sigmetrics_2024_lechowicz2024online} considers smoothed online conversion with a simple 
metric switching cost and a long-term constraint 
 $\sum_t^T{x_t}=1$ where $x_t\in\mathbb{R}^+$ is the scalar action at time $t$
(irrelevant to fairness). Besides different hitting and switching costs,
our work is substantially separated from \cite{SOCO_SimpleLongTermConstraint_UMass_AdamWierman_Sigmetrics_2024_lechowicz2024online}
as we consider a general and challenging fairness cost
that regularizes the long-term fairness of online actions
and necessitates our new algorithm \ouralg.

\textbf{Online optimization with fairness and long-term constraints.}
Our work 
is broadly relevant to online optimization with fairness
and long-term constraints \cite{Fair_Sequential_OptimalEnvyEfficieny_Sid_ChristinaYu_Cornell_OR_2023_10.1287/opre.2022.2397,OnlineOpt_Convex_LongTermConstraint_10.5555/2503308.2503322,Neely_Booklet}.
One key idea is to properly choose the Lagrangian multipliers
that correspond to long-term constraints or fairness. 
For example, some earlier works \cite{OCO_devanur2009adwords,OCO_feldman2010online} study online allocation problems by estimating a fixed Lagrangian multiplier using offline data,
while other works design online algorithms
by updating (or learning to predict) the Lagrangian multiplier in an online manner
\cite{devanur2019near,OCO_agrawal2014fast,OCO_zinkevich2003online,OCO_MULPLICATIVE_arora2012multiplicative,Shaolei_L2O_LAAU_OnlineBudget_AAAI_2023}. 
More recently, reinforcement learning has been applied to solve online problems
with long-term constraints \cite{L2O_NewDog_OldTrick_Google_ICLR_2019},
but it may not provide any worst-case performance guarantees in adversarial settings.

Importantly, some recent studies \cite{Shaolei_Replenishment_BudgetAllocation_SIGMETRICS_2024,OnlineOpt_EnergyHarvesting_Bregman_Neely_USC_SIgmetrics_2020_10.1145/3428337, OnlineAllocation_DualMirroDescent_Google_OperationalResearch_2022_doi:10.1287/opre.2021.2242, Fairness_OnlineAllocation_Regularized_Fairness_DMD_arXiv_2020_balseiro2021regularized} investigate online allocation problems in which the agent first fully observes the reward or cost function before selecting an action. 
In contrast, other recent work considers settings where certain components of the reward function are not revealed in advance \cite{fairness_unobserved_context_inria_neurips_2023, network_fair_dmd_bandit_feedback_chen_2024}, providing only bandit feedback or requiring additional cost for revelation. Such limitations preclude direct optimization of the reward via gradient-based methods.
Despite these differences, both types of settings can be addressed using dual mirror descent (DMD) or its variants, which update the Lagrangian multipliers based on the available context and the observed action sequence.
DMD can recover a family of classic online algorithms, including projected gradient descent \cite{OCO_zinkevich2003online,Neely_Booklet} and multiplicative weight \cite{OCO_MULPLICATIVE_arora2012multiplicative}. 
Similar algorithms are proposed for online stochastic optimization with distribution information in \cite{Online_stochastic_optimization_non_stationary_jiang2020online}.
While fairness can be converted into long-term constraints with
the introduction of an auxiliary variable,
it also substantially alters the algorithm designs and 
is known to be difficult to address,
as exemplified in the context of online budget allocation
\cite{Fairness_OnlineAllocation_Regularized_Fairness_DMD_arXiv_2020_balseiro2021regularized}.
More crucially, given the updated dual variable at each round, the objective function 
in these studies is \emph{independent} of the previous actions, but our objective function naturally has a memory due to  switching costs and hence requires a novel design of \ouralg.

%% file: tex/problem_definition.tex
\section{Problem Formulation}\label{sec:problem_formulation}

Consider a smoothed online optimization problem spanning an episode
of $T$ time rounds 
denoted by $[T]=\{1,\cdots,T\}$.
At each time $t\in[T]$, the agent observes a (potentially adversarially
chosen)
cost function $f_t(\cdot): \mathbb{R}^N\to\mathbb{R}^+$ and a matrix $A_t\in\mathcal{A}\subset \mathbb{R}^{M\times N}$, and chooses an
irrevocable action $x_t\in\mathcal{X}\subset \mathbb{R}^N$ without knowing the future information.
The agent incurs two costs at time $t$: a hitting cost $f_t(x_t)$ and a switching
cost $d(x_t, x_{t-1})$, which captures how well the current
cost $f_t(\cdot)$ and penalizes temporal changes
in actions, respectively. 
For notational convenience, we
define $\mathcal{Z}=\{Ax|A\in\mathcal{A},x\in\mathcal{X}\}$.
At each round $t$, the matrix $A_t$ maps
the agent's action $x_t$ into an $M$-dimension cost vector, with each dimension
corresponding to an entity for which \emph{fairness} needs to be taken into account. Thus, to regularize the online actions $x_t$ for fairness,  
we define a long-term horizon fairness cost 
 $g(\cdot): \mathbb{R}^M\to\mathbb{R}^+$ in terms of
$ \frac{1}{T}\sum_{t=1}^{T} A_t x_t$. 

The overall cost is denoted by ${cost}({x}_{1:T})$ and written as 
\begin{equation}\label{eqn:optimization_overall}
\begin{aligned}
     {cost}({x}_{1:T}) = \frac{1}{T}\Bigl( \sum_{t=1}^T f_t(x_t) + \sum_{t=1}^T d(x_t, x_{t-1}) \Bigr) + g \Bigl( \frac{1}{T}\sum_{t=1}^{T} A_t x_t \Bigr)
\end{aligned}
\end{equation}

\subsection{Assumptions}
We make the following standard assumptions as
in the prior literature \cite{SOCO_ROBD_Adam_NeurIPS_2019_10.5555/3454287.3454455,Neely_Booklet,SOCO_Memory_Adam_NIPS_2020_NEURIPS2020_ed46558a,SOCO_Revisiting_Nanjing_NIPS_2021_zhang2021revisiting}.

\begin{assumption}\label{assumption:model_cost}
    For each time $t\in[T]$, the hitting cost $f_t(x_t)$, switching cost $d(x_t, x_{t-1})$ and 
    long-term fairness cost $g( \frac{1}{T}\sum_{t=1}^{T} A_t x_t)$ satisfy  the following properties:

  $\bullet$  The hitting cost $f_t(x_t)$ is continuous and non-negative within the action set $X$. \\
  $\bullet$  The switching cost $d(x_t, x_{t-1})$ is the scaled squared $L_2$-norm, i.e., $d(x_t, x_{t-1}) =  \frac{\beta_1}{2} \| x_t - x_{t-1} \|^2$, where 
  $x_0$ is the initial action before the start of the episode. \\
 $\bullet$  The diameter of the vector set $\mathcal{Z}=\{Ax|A\in\mathcal{A},x\in\mathcal{X}\}$ is bounded
        by $Z$, i.e., 
    $\sup_{x_t, x_t' \in \mathcal{X}, A_t,  A_t' \in \mathcal{A}}\|A_t x_t- A_t' x_t'\|\leq Z$, where $\|\cdot\|$ is the $l_2$-norm unless otherwise noted. \\ 
$\bullet$   The fairness cost $g(y)$ is convex  
and $L$-Lipschitz, i.e., $g(y)-g(y')\leq L\cdot \|y-y'\|$ for any $y,y'\in\mathcal{Z}$. 
\end{assumption}

These assumptions, e.g., non-negativity, are needed for theoretical analysis
and widely considered in the literature \cite{SOCO_ROBD_Adam_NeurIPS_2019_10.5555/3454287.3454455,Neely_Universal}.
Additionally, it is common to consider a convex $L$-Lipschitz fairness cost. For example,
a well-known fairness cost that prioritizes the minimization
of higher costs for disadvantaged entities 
is the $l_p$ norm, i.e., $g(y)=\|y\|_p$ for $p\geq1$,
as considered in a variety of applications including fair federated learning
\cite{Fair_qFed_FederatedLearning_CMU_ICLR_2020_Li2020Fair}, budget
allocation 
\cite{Fairness_OnlineAllocation_Regularized_Fairness_DMD_arXiv_2020_balseiro2021regularized}, server scheduling \cite{BansalPruhs_STOC2003},
and geographical load balancing \cite{Shaolei_Equity_GLB_Environmental_AI_eEnergy_2024}, among many others. 
As a special case when $p\to\infty$, the $l_{\infty}$-norm  
fairness
cost addresses the classic minimax fairness.

In numerous online problems, including energy scheduling in power grids, object tracking in robotic-human interactions, server capacity provisioning in data centers, and more in \cite{SOCO_Memory_Adam_NIPS_2020_NEURIPS2020_ed46558a,SOCO_Revisiting_Nanjing_NIPS_2021_zhang2021revisiting,SOCO_ML_ChasingConvexBodiesFunction_Adam_COLT_2022}, both long-term cost and smoothness are crucial considerations.
To better motivate the consideration and make our model more concrete, we provide several application examples in Appendix~\ref{sec:real_world_application}.

%% file: tex/hardness_long_term.tex
\section{The Hardness of Long-term Fairness Regularizer}

Compared with switching cost which can be calculated with the action in the subsequent step, the long-term regularizer can be only evaluated at the end of the episode of $T$, with the complete action and context sequence. 
Thus, even intuitively, handling the long-term regularizer is considerably more challenging than managing switching costs. In this subsection, we will provide rigorous analysis on the impact of this inherent difficulty on two commonly considered performance metrics: regret and competitive ratio. 

Our theoretical results establish fundamental lower bounds on the performance gap between online and offline algorithms in the presence of such long-term dependencies, highlighting the hardness of the long-term fairness regularizer for online optimization even (in the absence
of the switching cost).

\begin{theorem}\label{thm:regret_linear_bound}
Assume that there is no switching cost (i.e., $\beta_1=0$).
Let $x_{1:T}^*$ denote the optimal offline action sequence.
For any online algorithm that produces an online sequence
of $x_{1:T}^\dagger$, the worst-case regret, defined as $\max_{(f_{1:T},A_{1:T})} \Bigl[ cost(x_{1:T}^\dagger) - cost(x_{1:T}^*) \Bigr]$, is lower-bounded by a positive constant, i.e. ${\Omega(1)}$. 
\end{theorem}

The proof idea of Theorem~\ref{thm:regret_linear_bound} is to construct adversarial future context sequence based on the historical actions of the online algorithm, which can be found in Appendix \ref{sec:regret_linear_bound_proof}. 
Notably, as the hitting and long-term costs are already averaged by the length of episode $T$, the constant performance gap in Theorem~\ref{thm:regret_linear_bound} can be translated to a linear regret when evaluating the total cost.
Furthermore, with the switching cost eliminated ($\beta=0$), this linear regret is solely attributed to the long-term cost.
The lower bound on the cost gap depends on the size of the action and context spaces, specifically quantified by the maximum norms of the action and context matrices (their diameters). Our proof establishes this lower bound by constructing an adversarial context sequence and identifying an offline optimal solution, both within a unit $l_2$ ball. If the diameters of these spaces further increase, the lower bound on the regret would also increase accordingly through a simple scaling of the adversarial sequence. Thus, if this diameter grows with the episode length $T$, the cost gap will similarly increase with $T$, potentially leading to superlinear regret in terms of the total cost. 

To conclude, the findings in Theorem~\ref{thm:regret_linear_bound} highlight the fundamental limitation imposed by the lack of future context information, preventing any online algorithm from asymptotically matching the offline optimal cost. 

\begin{theorem}\label{thm:cr_linear_bound}
Assume that there is no switching cost (i.e., $\beta_1=0$).
Let $x_{1:T}^*$ denote the optimal offline action sequence.
For any online algorithm that produces an online sequence
of $x_{1:T}^\dagger$,
the competitive ratio, defined as $C^\dagger = \max_{(f_{1:T},A_{1:T})} \frac{\text{cost}(x_{1:T}^\dagger)}{\text{cost}(x_{1:T}^*)}$, is lower-bounded by $\Omega(T)$, i.e., $C^\dagger \gtrsim \Omega(T)$.
\end{theorem}

In addition to regret, the competitive ratio is a widely used metric for evaluating online algorithms, particularly against a dynamic benchmark. Unlike a static benchmark, the dynamic setting allows the offline optimal action to vary over time, potentially achieving significantly lower costs. In such dynamic scenarios, achieving sublinear regret may be impossible, making the competitive ratio a more suitable performance measure. 
A finite competitive ratio guarantees that an online algorithm's cost is bounded by a constant multiple of the offline optimal cost. For problems with only switching and hitting costs, \cite{beyond_online_balanced_descent_goel2019beyond} achieved the state-of-the-art finite competitive ratio. 
Given that our problem setting also employs a similar dynamic benchmark, competitive ratio appears to be an appropriate evaluation metric. However, the inclusion of a long-term regularizer fundamentally alters this landscape.

The inherent difficulty introduced by such regularization is a long-standing research question. For example, \cite{Fairness_OnlineAllocation_Regularized_Fairness_DMD_arXiv_2020_balseiro2021regularized} considers a related online optimization problem focused on maximizing total reward under constraints on both the context sequence and the offline optimal benchmark. They conjecture that no online algorithm could achieve vanishing regret or a finite competitive ratio. Based on our Theorem \ref{thm:cr_linear_bound} and \ref{thm:regret_linear_bound}, we rigorously confirm that in this setting, neither a finite competitive ratio nor vanishing regret is attainable, even in the absence of switching costs. 
These theorems highlight the fundamental challenge introduced by the long-term regularizer. To enable meaningful performance comparisons despite this challenge, we slightly constrain the offline optimal benchmark and establish an asymptotic competitive ratio under this modified benchmark.

%% file: tex/performance.tex
% \subsection{Performance Metric}

% \pengfei{Will revise this part accordingly to highlight we have to constrain the offline optimal benchmark to some extent.}
% To measure the quality of an online algorithm in an adversarial setting, 
% the worst-case competitive ratio and regret are two common performance metrics, each having its distinct setting and applicable benchmark.
% While the (static) regret against a static optimal action
% is often considered in
% the online learning setting where the agent chooses an action before observing
% the cost function, the competitive ratio against a certain dynamic benchmark is the primary metric in the online optimization setting
% where the agent chooses an action after observing the cost function.

% \revise{
% While the worst-case competitive ratio and regret are two common performance metrics to measure the quality of an online algorithm in an adversarial setting, the fundamental challenge brought by the long-term regularizer make both of them linearly increasing with the episode length $T$ for any online algorithm. To provide a more meaningful comparison, we 
% define the following constrained benchmark, and evaluate the competitive ratio of an online algorithm asymptotically. }

\begin{definition}[$(R, \delta)$-optimal algorithm]\label{def:opt_R_delta}
   % Consider the setting where complete offline information is available, including the hitting cost $f_t(\cdot)$ and the time-varying coefficient $A_t$ over an episode of length $T$. 
    An action sequence $x_{1:T}^*$ is $(R, \delta)$-optimal if it satisfies: 
    %the following:
    \begin{equation}\label{eqn:R_delta_opt}
    \begin{aligned}
        & x_{1:T}^* = \arg \min_{x_{1:T}} {cost}({x}_{1:T}),   \;\;\;    s.t., \;  & \sum_{k=1}^K \biggl\| \sum_{t=(k-1)R+1}^{k R} A_t  x_t^*  - \frac{R}{T}\sum_{t=1}^{T} A_t  x_t^*  \biggr\| \leq \delta,
    \end{aligned}
    \end{equation}
    where  $R$ represents the frame size and $K$ is a positive integer denoting the number of frames in an episode such that $R\cdot K=T$, and
    $\delta\geq0$ measures the allowable total deviation between the frame-wise $\sum_{t=(k-1)R+1}^{k R} A_t  x_t^*$ and episode-wise average $\frac{R}{T}\sum_{t=1}^{T} A_t  x_t^*$ scaled by $R$. 
\end{definition}

The constrained benchmark, denoted $(R,\delta)$-OPT, imposes a restriction on the offline optimal solution or the adversarial context sequence. Specifically, the parameter $\delta$ limits the total deviation between the frame-wise average fairness vector, $A_t x_t^*$, and its episode-wise counterpart, scaled by the frame length $R$. The stringency of this $(R,\delta)$ constraint is inversely related to the values of $R$ and $\delta$; increasing either parameter makes the constraint less restrictive, and vice versa. This benchmark effectively limits the power of the offline optimum by assuming its total variance is sublinear with respect to the episode length $T$. If the total variance of the offline optimal action sequence were permitted to grow linearly with $T$, the environment can continuously shift the average of the optimal long-term cost. In other words, the offline optimal long-term cost is able to maintain a constant distance from the online algorithm's cost, thus explaining the constant cost gap observed in Theorem \ref{thm:regret_linear_bound}. Therefore, we assume $\delta$ in the offline optimal benchmark is always sublinear with respect to the episode length $T$.
Moreover, similar constrained offline optimal benchmarks have also been widely
considered in the literature, such as the 
switching-constrained OPT in smoothed online optimization \cite{SOCO_ROBD_Adam_NeurIPS_2019_10.5555/3454287.3454455} 
and frame-wise OPT with limited future information \cite{Neely_Booklet,Neely_Universal}.
Notably, in many practical applications such as data center scheduling, this is a very natural assumption, where the context $A_t$ exhibits some periodicity as evidenced by the diurnal workloads. In this case
even the actions chosen by the unconstrained offline algorithm can exhibit a periodic pattern, thus resulting in a sufficiently small $\delta$ for a certain large $R$. 

In the following, we interchangeably refer to
$(R,\delta)$-OPT as OPT whenever applicable without ambiguity. Next, we define the asymptotic competitive ratio \cite{OnlineAllocation_DualMirroDescent_Google_OperationalResearch_2022_doi:10.1287/opre.2021.2242} by considering
$T\to\infty$ as follows.

\begin{definition}[Asymptotic competitive ratio]\label{def:asymptotic_cr}
    Given an offline $(R, \delta)$-optimal algorithm OPT, an algorithm ALG is asymptotically $C$-competitive against OPT if for any
    problem instance $\gamma =\{f_t, A_t|t=1,
    \cdots, T\}$,  it satisfies
    $ \sup_{\gamma} \lim_{T \rightarrow \infty} \biggl[ cost(ALG) - C \cdot cost(OPT)  \biggr] \leq 0$.
%    \begin{equation}\label{eqn:cr_definition}
%        \sup_{\gamma} \lim_{T \rightarrow \infty} \biggl[ cost(ALG) - C \cdot cost(OPT)  \biggr] \leq 0.
%    \end{equation}
\end{definition}

%\shaolei{We don't need this. I think a fairness function is enough.}\pengfei{OK, I can absorb this to Assumption 2.2, I just need the diameter of the metric space}
%\begin{definition} [Metric space of fairness function]
%    Given the action set $\mathcal{X}$ and coefficient set $\mathcal{A}$, we define the set $\mathcal{Z} = \{ A \cdot x | A \in \mathcal{A}, x \in \mathcal{X} \}$ as the value of fairness function. The metric space of fairness function is defined as $(\mathcal{Z}, d)$, where metric $d$ is the $l_2$ norm.  
%\end{definition}

% The benchmark action sequence $x_{1:T}^*$ is the optimal solution for the following problem

% We define the  action sequence  $x_{1:T}^*$ as $(R, \delta)$-optimal, where $R$ is the frame length and $\delta$ is the total difference between the frame-wise averages and the episodic-wise average. 

%% file: tex/algorithm.tex
\section{\ouralg: Fairness-Regularized Online Balanced Descent}

We propose a novel online algorithm \ouralg
for fairness-regularized
smoothed online optimization defined in Eqn.~\eqref{eqn:optimization_overall}. 
Compared to the existing smoothed online optimization
literature without fairness consideration \cite{SOCO_Memory_Adam_NIPS_2020_NEURIPS2020_ed46558a,SOCO_ROBD_Adam_NeurIPS_2019_10.5555/3454287.3454455,SOCO_Memory_FeedbackDelay_Nonlinear_Adam_Sigmetrics_2022_10.1145/3508037,Shaolei_Learning_SOCO_Delay_NeurIPS_2023},
the crux of \ouralg is to decompose the horizon fairness
cost $g \bigl( \frac{1}{T}\sum_{t=1}^{T} A_t x_t \bigr)$ into online 
costs 
by introducing a new regularizer and an auxiliary variable that
carefully balances the hitting cost, switching cost, and fairness cost. 

\subsection{Algorithm Design}

Even in the absence of the fairness cost, the problem of smoothed
online optimization is already challenging due to the potential
conflicts between minimizing the current hitting cost $f_t(x_t)$
and staying closer to the previous action to reduce the switching cost 
$d(x_t,x_{t-1})$. The fairness cost $g \bigl( \frac{1}{T}\sum_{t=1}^{T} A_t x_t \bigr)$ is defined in terms of the long-term
average of $\frac{1}{T}\sum_{t=1}^{T} A_t x_t $ and hence
cannot be determined until the end of time $T$, further creating substantial challenges. \ouralg addresses these challenges by introducing an auxiliary
variable and decomposing
$g \bigl( \frac{1}{T}\sum_{t=1}^{T} A_t x_t \bigr)$ into an online version,
and by balancing the hitting cost, switching cost, and fairness cost.

\subsubsection{Decomposing the fairness cost}
To address the challenges stemming from the horizon fairness cost
$g \bigl( \frac{1}{T}\sum_{t=1}^{T} A_t x_t \bigr)$, 
we introduce an auxiliary variable $z_t\in\mathcal{Z}=\{z=Ax|A\in\mathcal{A},x\in\mathcal{X}\}$. 
In the long run, we aim to choose the actions that \emph{approximately} satisfy the condition $\sum_{t=1}^TA_tx_t = \sum_{t=1}^Tz_t$. This ensures that the horizon-wise average of the auxiliary variable can serve as an effective estimator for the horizon fairness cost.
Alternatively, we can view
$z_t$ as a dynamic ``budget'' that is allocated to the agent to guide its online actions. Thus, we reformulate the problem \eqref{eqn:optimization_overall}
as follows:
\begin{equation}
\label{eqn:optimization_overall_modified}
\begin{aligned}
    \min_{{x}_{1:T},z_{1:T}} &\frac{1}{T}\left( \sum_{t=1}^T f_t(x_t) + \sum_{t=1}^T d(x_t, x_{t-1}) \right) + \frac{1}{T}\sum_{t=1}^{T}g \bigl( z_t\bigr),\;\;\; s.t. &\; \sum_{t=1}^TA_tx_t = \sum_{t=1}^Tz_t,
\end{aligned}
\end{equation} 
whose optimal actions are denoted
by $(\hat{x}_{1:T}, \hat{z}_{1:T})$.
Importantly, the offline optimal cost of \eqref{eqn:optimization_overall_modified}
is the same as that of the original problem in \eqref{eqn:optimization_overall}.
This point can be seen by considering 
$\hat{z}_{1}=\cdots=\hat{z}_T=\frac{1}{T}\sum_{t=1}^TA_t \hat{x}_{t}$
and $\hat{x}_{1:T}={x}_{1:T}^*$ where ${x}_{1:T}^*$ is the offline
optimal solution to \eqref{eqn:optimization_overall}.

While the reformulated problem \eqref{eqn:optimization_overall_modified}
decomposes the horizon fairness cost $g \bigl( \frac{1}{T}\sum_{t=1}^{T} A_t x_t \bigr)$ by the introduction
of $z_{1:T}$ and has the same optimal cost as \eqref{eqn:optimization_overall},
it presents a new challenge due to the long-term constraint 
$\sum_{t=1}^TA_tx_t  =  \sum_{t=1}^Tz_t$ where $z_t$ itself is also an online action. 
To address this challenge, we note that unlike a real budget constraint that cannot be violated at any round \cite{OnlineAllocation_DualMirroDescent_Google_OperationalResearch_2022_doi:10.1287/opre.2021.2242},
 the long-term constraint in \eqref{eqn:optimization_overall_modified} only needs to be approximately
satisfied. Therefore, we relax the constraint and instead minimize the Lagrangian form of \eqref{eqn:optimization_overall_modified} as follows:
\begin{equation}
\label{eqn:optimization_overall_modified_lagrangian}
\begin{aligned}
    \min_{{x}_{1:T},z_{1:T}} & \frac{1}{T}\left( \sum_{t=1}^T f_t(x_t) + \sum_{t=1}^T d(x_t, x_{t-1}) \right) + \frac{1}{T}\sum_{t=1}^{T}g \left( z_t\right) + \kappa_t\left(\sum_{t=1}^TA_tx_t-\sum_{t=1}^Tz_t\right)
\end{aligned}
\end{equation}
where $\kappa_t\in\mathbb{R}^{M}$ is the estimated Lagrangian multiplier at round $t\in[T]$.

Had all the future information $\{f_{1:T}(\cdot), A_{1:T}\}$ been provided in advance, we could easily determine the optimal $\kappa_1=\cdots=\kappa_T=\kappa^*$. But, for online optimization,
we need to update $z_t$ based on the currently available information without knowing the future. Moreover, interestingly, given $\kappa_t$ at round $t$, $x_t$ and 
$z_t$ can be independently solved in \eqref{eqn:optimization_overall_modified_lagrangian}. 
Thus, we propose to employ mirror descent to update $\kappa_t$ online, based
 on which we subsequently optimize $x_t$ and $z_t$ at each round $t\in[T]$.

While \cite{OnlineAllocation_DualMirroDescent_Google_OperationalResearch_2022_doi:10.1287/opre.2021.2242,Fairness_OnlineAllocation_Regularized_Fairness_DMD_arXiv_2020_balseiro2021regularized,Neely_Booklet} use mirror descent to update the Lagrangian multiplier
for online allocation of a fixed given budget, our work has a crucial difference: given $\kappa_t$ at each round $t$, our objective includes
\emph{memory} in the form of a switching cost $d(x_t,x_{t-1})$,
whereas the objective in \cite{OnlineAllocation_DualMirroDescent_Google_OperationalResearch_2022_doi:10.1287/opre.2021.2242,Fairness_OnlineAllocation_Regularized_Fairness_DMD_arXiv_2020_balseiro2021regularized,Neely_Booklet} only depends on the current action $x_t$.
This difference voids the key steps in the proof techniques of 
\cite{OnlineAllocation_DualMirroDescent_Google_OperationalResearch_2022_doi:10.1287/opre.2021.2242,Fairness_OnlineAllocation_Regularized_Fairness_DMD_arXiv_2020_balseiro2021regularized,Neely_Booklet} and requires additional care to the switching cost.

\begin{figure}[t]
\vspace{-1em}
\begin{algorithm}[H]
   \caption{\ouralg: Fairness-regularized Online Balanced Descent}\label{alg:algorithm_main}
\begin{algorithmic}[1]
   \STATE {\bfseries Input:} Initial $\kappa_1$, reference  function $h(\cdot)$, and learning rate $\eta$ 
   \FOR{$t=1$ {\bfseries to} $T$}
   \STATE Receive $f_t(\cdot)$ and $A_t$
   \STATE Obtain the action $x_t$ and auxiliary variable $z_t$ by solving the following:
   \begin{equation}\label{eqn:single_opt_1}
    \begin{aligned}
        \min_{x_t \in \mathcal{X}_t, z_t \in \mathcal{Z}}  & \Bigl(f_t(x_t) + \lambda_1 d(x_t, x_{t-1}) + \frac{\lambda_2}{2} \| x_t - v_t \|^2  + \kappa_t \cdot A_t  x_t  + g(z_t)- \kappa_t z_t\Bigr),
    \end{aligned}
   \end{equation}
   where $v_t=\arg\min_{x\in\mathcal{X}}f_t (x)$.
   \STATE Obtain a stochastic subgradient of $\kappa_t$:
   $d_t = z_t - A_t\cdot x_t$
   \STATE Update the dual variable by mirror descent:
   $ \kappa_{t+1} = \arg\min_{\kappa \in \mathbb{R}^{M}} \langle d_t, \kappa \rangle + \frac{1}{\eta} V_h(\kappa, \kappa_t),$ 
    where $V_h(x,y)=h(x)-h(y)-\nabla h(y)^\top (x-y)$ is the Bregman divergence.
   \ENDFOR
\end{algorithmic}
\end{algorithm}
\vspace{-2em}
\end{figure}

\subsubsection{Balancing the hitting cost, switching cost, and fairness cost}

To further reconcile
conflicts between minimizing the current hitting cost $f_t(x_t)$
and staying closer to the previous action to reduce the switching cost 
$d(x_t,x_{t-1})$, we propose \ouralg that uses
the term $\kappa_t A_tx_t$ as a fairness regularizer while optimizing
$x_t$ online. The fairness regularization effect of $\kappa_t A_tx_t$
comes from the fact that it is intended to meet
the long-term constraint $\sum_{t=1}^TA_tx_t = \sum_{t=1}^Tz_t$ for fairness,
while ``balanced'' is due to two new hyperparameters $\lambda_1 \in (0,1]$ and $\lambda_2 \in [0,\infty)$ 
to balance the hitting cost $f_t(x_t)$ and switching cost $d(x_t,x_{t-1})$.
The hyperparameters $\lambda_1 \in (0,1]$ and $\lambda_2 \in [0,\infty)$  
can be optimally set to optimize the asymptotic competitive ratio in Theorem~\ref{thm:dmd_algorithm_switch}.
Furthermore, given $\kappa_t$, the auxiliary variable $z_t$ can be easily
optimized by solving $g(z_t)-\kappa_tz_t$.  

Before completing the design of \ouralg, we introduce
a reference function $h(\cdot):\mathbb{R}^{M}\to \mathbb{R}$
to define the Bregman divergence in our update of $\kappa_t$ using mirror descent.
Specifically, we choose $h(\cdot):\mathbb{R}^{M}\to \mathbb{R}$ that
is $l$-strongly convex and {$\beta_2$}-smooth 
\cite{MirrorDescent_SmoothReference_NeurIPS_2017_NIPS2017_b6e71087}.
For example,  Line~7 of Algorithm~\ref{alg:algorithm_main} follows the classic
additive update rule $\kappa_{t+1}=\left[\kappa_t-\eta d_t\right]^+$ where $\eta>0$
is the learning rate  when $h(\kappa)=\frac{\|\kappa\|^2}{2}$, whereas
it becomes the multiplicative update when $h(\kappa)=\kappa^T\log(\kappa)$
\cite{OnlineAllocation_DualMirroDescent_Google_OperationalResearch_2022_doi:10.1287/opre.2021.2242}. The specific choice
$h(\cdot)$ does not affect our performance analysis in Theorem~\ref{thm:dmd_algorithm_no_switch} and \ref{thm:dmd_algorithm_switch}.

Next, we formally describe \ouralg in Algorithm~\ref{alg:algorithm_main}.
In Line~4, we optimize $x_t$ to balance the hitting
cost, switching cost, and fairness cost. In Line~5, we optimize
$z_t$ as a fairness budget for guiding the optimization of $x_t$
in the future. In Lines~6 and~7, we update
$\kappa_{t+1}$ using mirror descent.

\subsection{Performance Analysis}\label{sec:analysis}

We present the analysis of \ouralg in terms of the asymptotic competitive ratio as follows. 
 
\begin{theorem}\label{thm:dmd_algorithm_no_switch}
Suppose that $x_{1:T}^*$ is the action produced by the $(R, \delta)$-optimal algorithm OPT. Assume that there is no switching cost (i.e., $\beta_1=0$). By setting the learning rate as
$\eta=\mathcal{O}( T^{-\frac{1}{3} })$ and parameters $\lambda_1 = \lambda_2 = 0$, and initializing $\kappa_1=\mathcal{O}(\frac{1}{{T}})$ (or $\kappa_1=0$
if $h(\kappa)=\frac{\|\kappa\|^2}{2}$),
the cost of \ouralg is upper bounded by
    \begin{equation}\label{eqn:thm_sublinear_bound}
    \begin{aligned}
        cost(x_{1:T}) 
        \leq&  cost(x_{1:T}^*) + 
    \biggl[\frac{\eta}{2l} Z^2 R +  Z \|\kappa_1 \| +  \beta_2 L \sqrt{\frac{1}{T}(\frac{Z^2}{l^2} + 2\frac{L Z}{\eta l} + \frac{2}{\eta^2 l} Z \| \kappa_1\|)} \biggr] + \frac{L\delta}{T}\\
    = & cost(x_{1:T}^*) +
    \mathcal{O}(T^{-\frac{1}{3}}) + \frac{L\delta}{T},
    \end{aligned}
\end{equation}
where the parameters $m$,  $Z$ and $L$ are specified
in Assumption~\ref{assumption:model_cost}, and $l$ and
 $\beta_2$ are the strong convexity and smoothness parameters of the reference function $h(\cdot)$, respectively.
\end{theorem} 

In Theorem~\ref{thm:dmd_algorithm_no_switch}, \ouralg achieves an asymptotic competitive ratio of 1 when the total deviation $\delta$ is sublinear in $T$. The total deviation $\delta$ quantifies the fundamental difficulty of the online problem itself, independent of the algorithm design. As discussed regarding the hardness of the long-term regularizer, if $\delta$ grows linearly with $T$, no online algorithm can attain sublinear regret.
Therefore, Theorem~\ref{thm:dmd_algorithm_no_switch} demonstrates that \ouralg achieves vanishing regret as the episode length approaches infinity, even in worst-case scenarios. This vanishing regret highlights the effectiveness of \ouralg in addressing the long-term regularizer, which is achieved by decomposing the long-term cost with a set of auxiliary variables. Furthermore, the dual variable within \ouralg dynamically adapts through continuous online updates, which drives the cost of \ouralg to eventually converge to the offline optimal benchmark.

\begin{theorem}\label{thm:dmd_algorithm_switch}
Suppose that $x_{1:T}^*$ is the action produced by the $(R, \delta)$-optimal algorithm OPT
and that the hyperparameters
 $\lambda_1 \in (0, 1]$ and $\lambda_2 \in [0, \infty)$. 
Assume further the hitting cost $f_t(\cdot)$ is $m$-strongly convex for any $t=1,\cdots,T$. 
By initializing {$\kappa_1=\mathcal{O}(\frac{1}{{T}})$ (or $\kappa_1=0$
if $h(\kappa)=\frac{\|\kappa\|^2}{2}$)} and setting the learning rate
$\eta=\mathcal{O}(T^{-\frac{1}{3}})$,
the cost of  \ouralg is upper bounded by
\begin{equation}\label{eqn:thm_cr_bound}
    \begin{aligned}
        cost(x_{1:T}) 
        \leq& C \cdot cost(x_{1:T}^*) + \frac{1}{\lambda_1}
    \Bigl[\frac{\eta}{2l} Z^2 R +  Z \|\kappa_1 \| +  \beta_2 L \sqrt{\frac{1}{T}(\frac{Z^2}{l^2} + 2\frac{L Z}{\eta l} + \frac{2}{\eta^2 l} Z \| \kappa_1\|)} + \frac{L\delta}{T}\Bigr]\\
    = & C \cdot cost(x_{1:T}^*) +
    \mathcal{O}(T^{-\frac{1}{3}}) + \frac{L\delta}{\lambda_1 T},
    \end{aligned}
\end{equation}
where $C = \max\{\frac{m+\lambda_2}{m\lambda_1}, 1 + \frac{\lambda_1 \beta_1}{m+\lambda_2}\}$. The parameters $m$,  $Z$ and $L$ are specified
in Assumption~\ref{assumption:model_cost}, and $l$ and
 $\beta_2$ are the strong convexity and smoothness parameters of the reference function $h(\cdot)$, respectively.
 In addition,
by considering $\delta$ in any sublinear forms in $T$ and optimally setting $\lambda_1 \in (0, 1]$ and $\lambda_2 \in [0, \infty)$
such that
$m+\lambda_2 = \frac{\lambda_1 m}{2}\left(1+\sqrt{1+\frac{4 \beta_1}{m}} \right)$, \ouralg achieves an asymptotic competitive ratio of $\frac{1}{2}\left( 1+ \sqrt{1+\frac{4 \beta_1 }{m}}\right)$ against  $(R, \delta)$-OPT when $T\to\infty$.
\end{theorem}

As demonstrated in Equation~\eqref{eqn:thm_cr_bound}, by properly setting the learning rate and initial Lagrange multiplier, \ouralg's cost asymptotically converges to within a constant factor of the $(R, \delta)$-constrained offline optimal cost. This convergence guarantees the stability of both the auxiliary variable and multiplier, ensuring robust performance irrespective of the context sequence. 
In our algorithm, the dual update is conducted using mirror descent, which generalizes the standard gradient descent as a special case when the reference function is chosen as $h(x)=\frac{1}{2}\|x\|^2$. Consequently, Theorem~\ref{thm:dmd_algorithm_switch} and Theorem~\ref{thm:dmd_algorithm_no_switch} also cover the performance analysis of dual updates with gradient descent by setting $l = \beta_2 = 1$. In addition, a comparison of these two theorems reveals that the constant $C$ in Theorem~\ref{thm:dmd_algorithm_switch} arises solely from the inclusion of the switching cost. 
Furthermore, by optimally setting $(\lambda_1, \lambda_2)$ to minimize $C$ in Equation~\eqref{eqn:thm_cr_bound}, \ouralg asymptotically achieves the competitive ratio of $\frac{1}{2}\Bigl( 1+ \sqrt{1+\frac{4{\beta_1}}{m}}\Bigr)$. This ratio matches the best-known competitive ratio lower bound \cite{SOCO_ROBD_Adam_NeurIPS_2019_10.5555/3454287.3454455} for the problem without a fairness cost. This observation highlights the superior performance of \ouralg and confirms that $C$ is inevitable due to the switching cost.

Unlike standard online smoothed optimization  \cite{beyond_online_balanced_descent_goel2019beyond}, the long-term fairness cost impacts the total cost of \ouralg in two-fold, primarily due to the $L$-Lipschitz property of the long-term fairness constraint. 
Generally speaking, larger $L$ implies faster change speed of fairness function, which introduces additional uncertainties when optimizing the auxiliary variables in $g(z_t)$. This imprecise estimation slows down the convergence speed of the dual variable, as reflected in the second last term in Eqn~\eqref{eqn:thm_cr_bound}. Moreover, since $\delta$ represents the discrepancy between the frame-wise and episode-wise averages of $A_t x_t$ in the offline optimal benchmark, a larger $L$ amplifies this difference in \ouralg's final fairness cost, as demonstrated by the last term in Equation~\eqref{eqn:thm_cr_bound}.
Thus, the overall performance of \ouralg is significantly influenced by the Lipschitz constant $L$, which bounds the first-order smoothness of the fairness cost.

The additive gap observed in Theorem~\ref{thm:dmd_algorithm_no_switch} and Theorem~\ref{thm:dmd_algorithm_switch} is directly influenced by the inherent difficulty associated with the long-term cost. This difficulty is captured by $R$ and $\delta$: as $R$ and/or $\delta$ increase, the potential for adversarial context sequences in the long-term cost becomes greater, which is a fundamental challenge for any online algorithm. Thus, $R$ and $\delta$ quantify the maximum level of adversarialness in the context sequence. However, many real-world applications are less adversarial, exhibiting predictable periodic patterns (such as the diurnal cycles in AI workload distributions). These patterns significantly reduce $\delta$, the long-term deviations. Consequently, for a reasonably large yet finite $R$, the total deviation between the online performance and the unconstrained offline optimal benchmark can remain constant or grow sublinearly with $T$.

In addition to the rigorous insights provided by Theorem~\ref{thm:dmd_algorithm_switch}, our proof technique also represents a novel technical contribution. Unlike existing %primal-dual 
proof techniques that typically rely on the assumption of 
independence between per-round objective functions, our approach does not require such an assumption. 
Removing the assumption of temporal independence is crucial in our fairness-regularized smoothed online optimization problem, as the switching cost functions inherently violate this assumption by incorporating historical actions. Consequently, comparing the costs of per-round objective functions between \ouralg and the offline optimal benchmark becomes highly challenging, as their cost functions are no longer identical under these conditions. 
To address this limitation, Lemma~\ref{lemma:total_cost} and~\ref{lemma:intermediate_cost} establish a theoretical foundation for such a comparison by introducing an intermediate benchmark with per-round objective functions aligned with those of \ouralg. 
This intermediate benchmark effectively bridges the gap between the two policies, each associated with distinct sets of cost functions, enabling a meaningful and rigorous comparison. 
Though the current result relies on the properties of squared $l_2$-norm in switching cost, 
it is promising to relax this assumption to accommodate more general strongly convex and smooth functions. Specifically, in Lemma~\ref{lemma:total_cost} and Lemma~\ref{lemma:intermediate_cost}, the results can be directly extended under strong convexity, since the squared $l_2$-norm is a special case where convexity and smoothness coincide. Furthermore, when converting the intermediate benchmark to the offline optimal benchmark, the gap between the two can also be bounded by the strong convexity of the switching cost.
It is interesting future work to study this more general setting. More broadly, the proof techniques in our approach represent a substantial development in addressing states influenced by past actions, extending beyond the limitations of existing approaches that are restricted to handling memoryless functions.

%% file: tex/simulation.tex
\section{Empirical Results}

 Our empirical study investigates the problem of fair resource provisioning for AI inference services in geographically separated data centers, where decisions incur instantaneous hitting costs (energy consumption and workload‐imbalance penalties), switching costs (reconfiguration overhead), and a long‐term fairness cost quantifying regional public‐health impacts from electricity generation pollution. \ouralg decomposes and dynamically optimizes all three cost components via auxiliary variables and mirror‐descent updates to improve cost efficiency, action smoothness, and long-term fairness simultaneously. In our simulation, we use the one‐week trace of normalized LLM inference requests from \cite{AzurePublicDataset}, and route demand across seven selected data centers (Arizona, Iowa, Illinois, Texas, Virginia, Washington, Wyoming) with publicly available electricity prices, PUE values, and health‐damage rates from WattTime \cite{watttime_dataset}. We compare \ouralg against five benchmarks: the offline optimal (\opt), offline fairness‐only (\fairopt), hitting‐cost minimizer (\hitmin), Regularized Online Balanced Descent (\robd), and Dual Mirror Descent (\dmd). 

We also test \ouralg with different learning rates and different weights for the fairness cost to test its robustness under different settings, with more details in Appendix~\ref{sec:appendix_experiment}. 
Our numerical results (Table~\ref{table:main_results} in Appendix~\ref{sec:appendix_experiment}) demonstrate that \ouralg achieves the lowest total cost among online methods, with the minimal cost gap to the offline optimal. Moreover, \ouralg attains the smallest fairness cost among online baselines, while maintaining low switching overhead. These results confirm that explicit incorporation and dynamic optimization of long‐term fairness in \ouralg yields superior performance across all cost dimensions.
 

%% file: tex/conclusion.tex
\section{Conclusion}\label{sec:conclusion}

We study a novel and challenging problem of fairness-regularized smoothed online convex optimization.
Our main contribution is the proposal of
 \ouralg to reconcile the tension between minimizing the hitting
cost, switching cost, and fairness cost.
Importantly, \ouralg offers a guaranteed worst-case asymptotic competitive
ratio against $(R,\delta)$-OPT 
for finite $R$ and sublinearly growing $\delta$
by considering the problem episode length $T\to\infty$.
Moreover, our analysis can recover the best-known results
for two existing settings (i.e., smoothed online optimization without fairness
and fairness-regularized online optimization without switching costs) as
special cases.  
Finally,  we run trace-driven experiments of dynamic computing resource provisioning for socially responsible AI inference to demonstrate
the empirical advantages of \ouralg over existing baseline solutions.

\section{Limitation and Impact Statements}
While simultaneously addressing fairness and action smoothness in online optimization, \ouralg is designed based on a set of assumptions, including non-negative hitting costs and squared $l_2$-norm switching costs. Therefore, it is interesting future work to overcome these limitations by, e.g., relaxing the strong convexity assumption, considering more general forms of switching costs, and/or incorporating potentially untrusted predictions of future information to further reduce the overall cost. Although our work can potentially increase awareness of fairness in  online optimization, we do not foresee  negative societal impacts or the need for safeguards due to the theoretical nature of our work.

\section*{Acknowledgment}
Pengfei Li, Yuelin Han, and Shaolei Ren were supported in part by the NSF under grants CNS-2007115 and CCF-2324941. Adam Wierman was supported by NSF grants CCF-2326609, CNS-2146814, CPS-2136197, CNS-2106403, and NGSDI-2105648 as well as funding from the Resnick Sustainability Institute.

%% file: tex/appendix_application.tex
\section{Real-world Applications with Long-term Costs}\label{sec:real_world_application}

In this section, we present several real-world application that emphasizes long-term fairness along with the switching cost. We provide concrete modeling for these applications and demonstrate how these two applications align closely with our problem formulation.

\subsection{Geographical Load Balancing with Balanced Public Health Costs}
The exponentially growing demand for AI has been driving
the recent surge of data centers around the globe, which are notoriously
energy-consuming and thus have huge environmental impacts in terms of
the local air pollution and carbon emissions (e.g. for fossil-fuel based power plants). 
The escalating air pollution resulting from AI computing significantly contributes to various health issues, including lung cancer, heart disease, and cardiovascular problems. Thus, the rise in ambient air pollution leads to substantial public health costs, encompassing lost workdays, increased medication usage, and increased hospitalization rates.
More critically, the public health costs of data centers are becoming increasingly unevenly distributed across different regions, disproportionately affecting marginalized communities.
To mitigate the uneven distribution of public health impacts from AI computation, an effective approach
is to leverage fairness-aware geographical
load balancing to fairly distribute the environmental costs across different data center locations \cite{Shaolei_Equity_GLB_Environmental_AI_eEnergy_2024}, and, consequently, public health risks across diverse data center locations.

Specifically, let $x_t\in\mathbb{R}^N_+$ be the number of active
servers or computing capacity to process incoming workloads in $N$ different data centers at time $t$. 
With a provisioning capacity of $x_t$, there is an operational cost that includes the energy cost and a penalty term for workload imbalance,
which can be modeled as the hitting cost $f_t(x_t)$ \cite{SOCO_DynamicRightSizing_Adam_Infocom_2011_LinWiermanAndrewThereska}.
When adjusting computing capacities, a switching cost $d(x_t,x_{t-1})$ arises due to server reallocation, workload migration, or wear-and-tear. Additionally, a health cost $A_t x_t$ is incurred across $N$ locations due to air pollution from electricity generation \citenewcite{AI_health_yuelin_2024}, influenced by the local grid's real-time fossil fuel intensity.
Further modeling details can be found in Appendix~\ref{appendix:simulation_setup}.

\subsection{Socially-fair AI Resource Provisioning} 
Large AI models often have multiple sizes, each with a distinct performance and resource usage tradeoff.
For example, the GPT-3 model family has eight different sizes, ranging
from the smallest one with 125 million parameters to the largest one with 175 billion parameters \citenewcite{ML_GPT3_Energy_Others_NIPS_2020_NEURIPS2020_1457c0d6}.
As a result, the amount of computing resources 
can directly affect the model selection and inference quality during inference,
and needs to be carefully determined based on the 
number of user requests from different regions. As
we become increasingly reliant on AI for acquiring
knowledge, insufficient resource provisioning can lead
to subpar AI model performance and could disproportionately jeopardize
the prospects of users/individuals from certain disadvantaged regions.
 Thus, the computing resource provisioning for users from different regions
must be fair in the long run. In addition, there
is a switching cost when adjusting the computing resource capacity (e.g., by activating
otherwise hibernating servers).

We consider a discrete-time model for the AI server provisioning problem, where $y_t \in \mathbb{R}^M$ represents the user demand from $M$ different regions at time $t$ and $x_t \in [x_{\min}, x_{\max}]$ denotes the amount of provisioned AI servers to meet the demand. 
The hitting cost $f(x_t, y_t)$ encompasses both the electricity expenses of the AI servers and a penalty term for any dropped workload demand.
The switching cost $d(x_t, x_{t-1})$ models the cost incurred by hardware and software reconfiguration, particularly when server provisioning undergoes frequent or substantial changes. 
At each time slot, we assume the provisioned AI servers are shared by each region in proportion to their user workload demand. To promote fair distribution of AI resources among the $M$ regions within the horizon, we account for each region's dynamic resource share, which varies based on its fluctuating workload demands.  
Specifically, we employ a min-max fairness function over the total amount of allocated AI resources for the $M$ regions.

%% file: tex/appendix_simulation.tex
\section{Experiment}\label{sec:appendix_experiment}

In this section, we run experiments of fair computing resource provisioning for AI inference services. Our results demonstrate that \ouralg can effectively reduce
the total fairness-regularized cost, highlighting the empirical advantages of \ouralg compared
to existing baseline solutions.

\subsection{Simulation Setup}\label{appendix:simulation_setup}
 
We investigate the challenge of fair resource provisioning for AI data centers that deliver inference services to dynamically routed AI workloads. At each time step, we dynamically route incoming user demand across geographically dispersed AI data centers and provision server/GPU resources accordingly.
This provisioning process incurs a hitting cost (e.g., energy cost, carbon emissions, and penalties for imbalanced workload distribution), as well as a switching cost representing the reconfiguration overhead associated with changes in provisioning decisions.
Furthermore, electricity consumption contributes to public health risks within the local community due to air pollution generated during electricity production. We incorporate a long-term fairness cost that quantifies the degree of public health impacts on the most affected region across all AI data centers.

\subsubsection{Model}
We consider a discrete-time model where each round (or time slot)
is one hour. As each round has the same duration, we interchangeably
use power and energy wherever applicable without ambiguity. 
To fulfill the demand for AI inference service at time $t$, denoted as $w_t$, 
we need to determine AI server/GPU provisioning action $x_t$. The resource provisioning decision is defined as $x_t = [x_{1,t}, \cdots, x_{N,t}]$, where $x_{i,t}$ denotes the provisioned server/GPU capacity for data center $i\in [N]$, subject to constraint $x_{i,t}\leq M_i$ where $M_i$ is its maximum capacity. 
Additionally, to guarantee the optimal quality of experience for users, we consider all the AI workloads are instantaneously served without further delays, where $\sum_{i=1}^N x_{i,t} = w_t$. 
The provisioning decision incurs a hitting cost (e.g., energy cost and penalty for imbalanced workload distribution), and a switching cost due to changing the provisioning decisions.
Furthermore, electricity consumption contributes to public health risks within the local community due to air pollution generated during electricity production. We incorporate a long-term fairness cost that quantifies the degree of public health impacts on the most affected region across all AI data centers. We model these three costs as follows.

The hitting cost encompasses the cost of energy consumption and penalties for imbalanced workload scheduling. Specifically, we model server power consumption as a linear function of provisioned computing resources, $q\cdot x_t$. And the non-IT power consumption (e.g., cooling systems) is modeled by multiplying the power usage efficiency (PUE) $\gamma_i$ over the power consumption at data center $i \in [N]$. 
Thus, the total electricity expenses of AI computing can be expressed as $\sum_{i=1}^N p_{i,t}^e \cdot \gamma_i \cdot q \cdot x_{i,t}$, where $p_{i,t}^e$ denotes the time-varying electricity price at location $i$.   
Additionally, given the fluctuating AI workload, it's highly imperative to achieve a balanced workload distribution, both spatially and temporally. Within time slot $t$, as the entire AI workload must be fulfilled collaboratively by $N$ data centers, imbalanced workload distribution can overload several data centers and result in low utilization rate for some data centers. The low utilization may leads to wasted reserved capacity and potentially increased idle power consumption. Additionally, from the perspective of whole horizon $T$, a smaller temporal variation in the provisioned server capacities is also preferred, considering the same amount of total workload. To this end, we incorporate a regularizer $u_1 \|x_{i,t}\|^2$ into the hitting cost of data center $i$. When it's summed over the entire planning horizon $T$, this term represents the squared $L_2$-norm of the provisioned server capacity vector, $[x_{i,1},\cdots, x_{i,T}]$.

To sum up, the  hitting cost at time $t$ is defined as
\begin{equation}
    f_t(x_t) = \sum_{i=1}^N p_{i,t}^e \cdot \gamma_i \cdot q \cdot x_{i,t} + u_1 \sum_{i=1}^N \| x_{i,t}\|^2 
\end{equation}
Besides, the switching cost $d(x_t, x_{t-1}) = \frac{u_2}{2}\|x_t-x_{t-1}\|^2$ penalizes the frequent or large changes between consecutive actions. Here, $u_2$ is a hyperparameter that reflects the overhead associated with hardware reconfiguration and communication arising from dynamically routed workloads.

\begin{table}
 \centering
 \footnotesize
 \begin{tabular}{c|c|c|c} 
 \toprule
 \textbf{Location} & \begin{tabular}[c]{@{}c@{}}\textbf{Average Health }\\\textbf{Price} (\$/MWh)\end{tabular} & \begin{tabular}[c]{@{}c@{}}\textbf{Average Electricity }\\\textbf{Price }(\$/MWh)\end{tabular} & \textbf{PUE ($\gamma_i$)} \\ 
 \hline
 Arizona & 17.29 & 77.7 & 1.18 \\ 
 \hline
 Iowa & 62.81 & 62.6 & 1.16 \\ 
 \hline
 Illinois & 49.93 & 82.6 & 1.35 \\ 
 \hline
 Texas & 48.19 & 63.0 & 1.28 \\ 
 \hline
 Virginia & 52.68 & 87.0 & 1.14 \\ 
 \hline
 Washington & 17.55 & 62.0 & 1.15 \\ 
 \hline
 Wyoming & 34.79 & 76.1 & 1.11 \\
 \bottomrule
 \end{tabular}
 \vspace{0.2cm}
 \caption{Information for selected data centers}
 \label{tab:info_dc}
 \vspace{-0.3cm}
\end{table}

Scaling laws are observed in large AI services \citenewcite{scaling_law_AI_kaplan2020}, such as large language models, demonstrating an exponential increase in computational requirements to achieve continuous performance improvements. This rapid escalation in computation translates to substantial energy consumption. Due to the air pollution generated during electricity production, the increased electricity consumption associated with these models contributes to public health risks within local communities.
Following the methodology in \cite{watttime_dataset}, we denote the health damage rate as $A_t = [A_{1,t}, \cdots, A_{N,t}]$, which reflects the associated air pollution and consequent health risks resulting from the electricity consumption.  
By strategically aligning AI computation with time periods and locations that are abundant with cleaner energy sources (e.g., solar, wind), the health damage rate can be significantly minimized.
At the same time, ensuring fair distribution of public health risks is crucial when dynamically routing AI workloads. We incorporate a long-term fairness cost that quantifies the degree of public health impacts on the most affected region across all AI data centers, formulated as below
\begin{equation}
    \begin{aligned}
        g \Bigl( x_{1:T}\Bigr) =  u_3 \biggl\| \frac{q}{T}  \sum_{t=1}^T A_t \cdot x_t^{\top} \biggr\|_p
    \end{aligned}
\end{equation}
where the $L_p$-norm can recover the widely-considered min-max fairness function as $p$ approaches infinity. 
By regularizing the overall public health risk in the long-term, we can not only promote social fairness and mitigate widening socioeconomic gaps, while maintaining sufficient flexibility for dynamic AI workload routing.

By summing up the hitting, switching and long-term fairness cost, 
the objective function is defined as below

\begin{equation}\label{eqn:cost_total_simulation}
    \begin{aligned}
         & \arg\min_{x_{1:T}}  \frac{1}{T} \sum_{t=1}^T f_t(x_t) + \frac{1}{T}\sum_{t=1}^T d(x_t, x_{t-1})  + g \Bigl( x_{1:T}\Bigr)\\
         s.t. & \;\;\;\;\; \sum_{i=1}^N x_{i,t} = w_t, \;\; \forall t \in [1,T]\\
         & \;\;\;\;\;\quad\quad x_{i,t} \leq M_i, \; \forall i \in [1,N], \; \forall t \in [1,T]
    \end{aligned}
\end{equation}

\subsubsection{Datasets and parameter settings} 
We employ a publicly available inference trace dataset for LLM services on Azure \cite{AzurePublicDataset}. We normalize a sample of coding-related inference requests processed by multiple LLM services within Azure.
These traces are collected between May 10th and May 16th, 2024. The dataset provides user demand patterns across different times of the week. 

Specifically, we aggregate the number of requests received hourly to simulate the total computational workload across different times of the day. 

To meet this workload, we assumed that the total computing demand could be distributed across seven selected data centers located in Arizona, Iowa, Illinois, Texas, Virginia, Washington, and Wyoming ($N$ = 7). The data center information is obtained from \citenewcite{MicrosoftLocalCommunities}. We use the average state industrial electricity price \citenewcite{EIA_OPEN_DATA}, denoted as $p_{i, t}^e$,  to calculate the electricity cost for each data center. 
For $\gamma_i$, we use the values reported by \citenewcite{MicrosoftDataCentersSustainability_PUE} to conduct our experiment.

For health impact analysis, air pollutant emissions resulting from electricity consumption contribute to various adverse effects, such as increased hospitalizations, higher medication usage, more frequent emergency room visits, and additional lost school days. 
These impacts can be quantified in economic terms using epidemiological and economic research on the associated health outcomes, collectively referred to as health costs. WattTime \cite{watttime_dataset} divides the United States into distinct regions and provides real-time health prices (\$/MWh), representing the health cost (\$) per unit of energy consumption (MWh) across different regions and time periods.
To determine the region in which each data center is located, we utilize the information from \citenewcite{MicrosoftLocalCommunities}. We use the health damage rate to estimate the health cost of energy consumption for each data center. Specifically, we set $A_t = [\gamma_i \cdot p_{i,t}^h]_{i=1}^{N}$, where $p_{i,t}^h$ represents the hourly average of WattTime’s health price.
Our online algorithm's ability to dynamically update its dual variable and learn from the online context sequence eliminates the need for prior training on a separate dataset. Therefore, we only build a testing dataset for all baseline algorithms. This testing dataset consists of 97 three-day (72-hour) context sequences, created by applying a sliding window across a one-week (168-hour) context sequence.

To provide readers with a high-level overview of the data centers' operational characteristics, Table \ref{tab:info_dc} summarizes several key metrics including the average electricity price and average health price observed from May 10th to May 16th, 2024, as well as the Power Usage Effectiveness (PUE) and geographic location for each facility.

In the default setting, we set $q = 1$ to map the provisioned computing resource to energy consumption. The weights are $u_1 = 10$ for the regularizer in hitting cost, $u_2 = 1000$ for switching cost and $u_3 = 3.5$ for the long-term fairness cost.
We choose the identical maximum capacity for each data center with $M_i = 1$ and normalize maximum workload traces according to the maximum capacity.
We use $p = \infty$ for the $l_p$ norm. These values are chosen to ensure that the hitting cost, switching cost, and fairness cost have comparable magnitudes. We set the initial dual variable $\kappa_1 = [3]_{i=1}^{N}$, $\lambda_1$ = 1, $\lambda_2$ = 30 for \dmd and \ouralg.

The hyperparameters $\lambda_1$ and $\lambda_2$ for \robd 
are optimally selected based on Theorem 4 in  \cite{SOCO_ROBD_Adam_NeurIPS_2019_10.5555/3454287.3454455}. For \dmd, we use the same set of hyperparameter as \ouralg, such as the default learning rate $\eta = 10^{-3}$, except that \dmd ignores the switching cost. Regardless of the algorithm. The total costs of all the baseline algorithms are calculated with Eqn~\eqref{eqn:cost_total_simulation}.

Our experiments are conducted on a MacBook Air with an M3 chip and 16 GB of memory. The average execution time per online algorithm over 72 time slots is approximately 1 second.

{
\renewcommand{\arraystretch}{1.05}
\begin{table*}[!t]
\scriptsize
\centering
\begin{tabular}{l|c|c|c|c|c|c|c|c} 
\toprule
\multirow{2}{*}{Metrics} & \multirow{2}{*}{\textbf{\opt}} & 
\multirow{2}{*}{\textbf{\fairopt}} & \multirow{2}{*}{\textbf{\hitmin}} & \multirow{2}{*}{\textbf{\robd}} & \multirow{2}{*}{\textbf{\dmd}} & \multicolumn{3}{c}{\textbf{\ouralg}}   \\ 
\cline{7-9}
&&&&& &$\eta = 10^{-2}$& $\eta = 10^{-3}$ & $\eta = 10^{-4}$   \\ 
\hline
Hitting Cost &171.30&177.20&\textbf{159.75}&163.63&169.46&170.06&167.80&167.47\\ 
\hline
Switching Cost &23.27&351.48&43.75&\textbf{23.16}&93.53&25.65&27.52&28.17\\
\hline
Fairness Cost & 36.36&33.33&140.12&111.85&54.16&\textbf{41.36}&51.05 &52.68 
  \\   
\hline
\textbf{Total Cost}&230.93&562.00&343.62&298.64&317.15&\textbf{237.07}&246.36&248.31   
\\
\bottomrule
\end{tabular}
\caption{The average costs of different algorithms in the default setting (i.e. $u_1 = 10$, $u_2 = 1000$ and $u_3 = 3.5$). 
Minimum costs for the online algorithms are highlighted in bold.   
}\label{table:main_results}
\vspace{-1.0em}
\end{table*}
\renewcommand{\arraystretch}{1.0}
}

\begin{figure}[t]
    \centering
    \includegraphics[width=\linewidth]{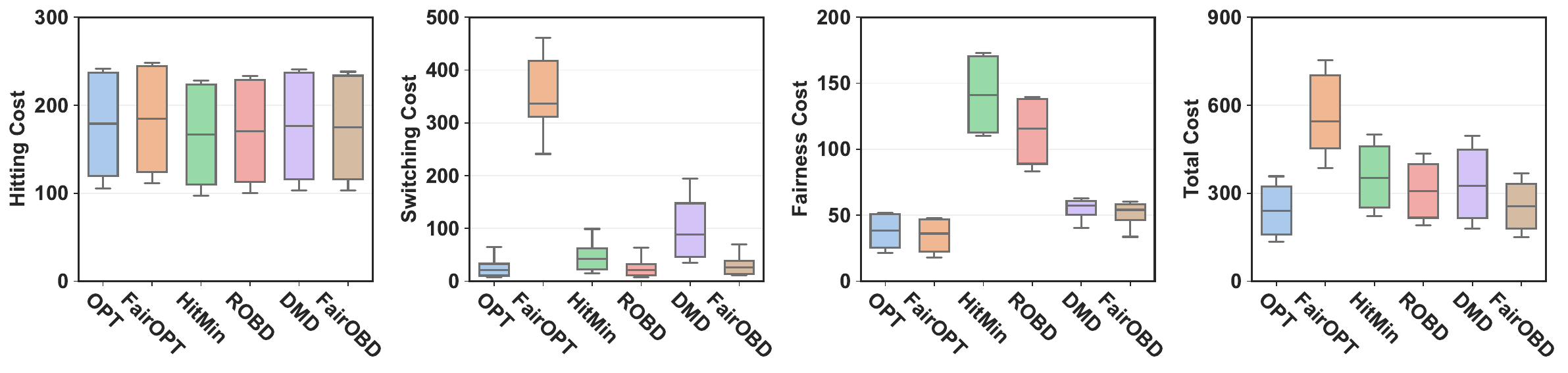}
    \vspace{-1.0em}
    \caption{Cost distributions of different algorithms in the default setting (i.e. $u_1 = 10$, $u_2 = 1000$ and $u_3 = 3.5$) 
    }
    \vspace{-1.0em}
    \label{fig:error-bar}
\end{figure}

\subsubsection{Baseline Algorithms}

We consider the following baseline algorithms for comparison.
\begin{itemize}
    \item \textbf{Optimal Fairness Offline (\fairopt)}: the offline optimal policy solely minimizing long-term fairness cost.
    \item \textbf{Hitting Cost Minimizer (\hitmin)}: the online algorithm chasing the minimizer of the hitting cost. 
    \item \textbf{Regularized Online Balanced Descent (\robd)}: the state-of-the-art online algorithm without accounting the long-term fairness cost.  
    \item \textbf{Dual Mirror Descent (\dmd)}: the online algorithm solely optimizing long-term fairness. 
    \item \textbf{Optimal Offline (\opt)}: the strongest optimal offline benchmark, corresponding to $(T,0)$-OPT in our analysis. No online algorithms can possibly outperform OPT.
\end{itemize}

Among these algorithms, \robd achieves the lowest competitive ratio for smoothed online convex optimization \cite{SOCO_ROBD_Adam_NeurIPS_2019_10.5555/3454287.3454455}, which is an adapted version of Eqn~\eqref{eqn:optimization_overall} by removing the fairness cost. Besides,  \dmd is proposed for online allocation problems without switching
costs \cite{Fairness_OnlineAllocation_Regularized_Fairness_DMD_arXiv_2020_balseiro2021regularized}, 
\citenewcite{Fair_OnlineAllocation_Regularized_Fairness_DMD_Google_ICML_2021_pmlr-v139-balseiro21a},
\cite{Neely_Universal}, 
which updates the dual variable $\kappa_t$ based on mirror descent techniques. 
We adapt \dmd with a reference function $h(\kappa)=\frac{\|\kappa\|^2}{2}$ in our experiment.

\subsection{Results}
We summarize the average costs of the online algorithms and their offline optimal benchmark under the default setting in Table~\ref{table:main_results}.
The proposed algorithm, \ouralg,  maintains robust performance across different learning rates $\eta$ and consistently outperforms other online baseline algorithms (\hitmin, \robd, and \dmd) in terms of total cost. 
Moreover, \ouralg attains the minimum fairness cost among all online baselines, demonstrating its effectiveness in minimizing the long-term fairness cost.

To illustrate the minimal achievable long-term fairness cost, we incorporate \fairopt as the offline fairness-optimal baseline, considering access to the complete context sequence. 
Due to the inherent trade-off between operational cost and fairness cost, the locations with the lowest health impact do not necessarily align with those offering lower electricity prices. Furthermore, exclusively targeting such locations would exacerbate fluctuations in server provisioning across different regions. These two factors explain why \fairopt achieves the lowest fairness cost, despite its hitting and switching costs being among the highest observed, as shown in Figure~\ref{fig:error-bar}. 
Conversely, \hitmin operates in the opposite direction, focusing exclusively on minimizing hitting cost while significantly increasing the other two costs. To address this challenge, \dmd resolves the tension between long-term fairness cost and instantaneous operational cost by decomposing the long-term costs and optimizing them using mirror descent techniques. This approach enables an effective balance between these two competing metrics.

\begin{table*}[t]
\scriptsize
\centering
\begin{tabular}{l|c|c|c|c|c|c|c|c} 
\toprule
\multirow{2}{*}{Metrics} & \multirow{2}{*}{\textbf{\opt}} & 
\multirow{2}{*}{\textbf{\fairopt}} & \multirow{2}{*}{\textbf{\hitmin}} & \multirow{2}{*}{\textbf{\robd}} & \multirow{2}{*}{\textbf{\dmd}} & \multicolumn{3}{c}{\textbf{\ouralg}}   \\ 
\cline{7-9}
&&&&& &$\eta = 10^{-2}$& $\eta = 10^{-3}$ & $\eta = 10^{-4}$   \\ 
\hline
Hitting Cost  & 168.88&177.20&\textbf{159.75}&163.63&169.46 &169.39&167.80&167.47\\ 
\hline
Switching Cost &22.92&351.49&43.75&\textbf{23.16}&93.53&25.67&27.52&28.17\\ 
\hline
Fairness Cost &20.33&16.67&70.06&55.93&27.08&\textbf{21.35}&25.52&26.34\\   
\hline
\textbf{Total Cost}  &212.13&545.35&273.56&242.71&290.07&\textbf{216.41}&220.84&221.97 \\ 
\bottomrule
\end{tabular}
\caption{{Cost comparison of different algorithms with the  fairness
weight $u_3 = 1.75$.}}
\label{table:results_small_u}
\end{table*}

\begin{table*}[t]
\scriptsize
\centering
\begin{tabular}{l|c|c|c|c|c|c|c|c} 
\toprule
\multirow{2}{*}{Metrics} & \multirow{2}{*}{\textbf{\opt}} & 
\multirow{2}{*}{\textbf{\fairopt}} & \multirow{2}{*}{\textbf{\hitmin}} & \multirow{2}{*}{\textbf{\robd}} & \multirow{2}{*}{\textbf{\dmd}} & \multicolumn{3}{c}{\textbf{\ouralg}}   \\ 
\cline{7-9}
&&&&& &$\eta = 10^{-2}$& $\eta = 10^{-3}$ & $\eta = 10^{-4}$  \\ 
\hline
Hitting Cost  & 171.81&177.20&\textbf{159.75}&163.63  &169.46&170.52&167.80&167.47\\ 
\hline
Switching Cost &23.97&351.49&43.75&\textbf{23.16}&93.53&25.70&27.52&28.17\\ 
\hline
Fairness Cost  &70.82&66.66&280.24&223.70&108.32&\textbf{82.40} &102.10&105.35 \\   
\hline
\textbf{Total Cost}  &266.60&595.35&483.75&410.49&371.31 &     \textbf{278.62}&297.41&300.99
\\ 

\bottomrule
\end{tabular}
\caption{{Cost comparison of different algorithms with the  fairness
weight $u_3 = 7$.}}\label{table:results_large_u}
\end{table*}

Without explicitly accounting for smoothness during the load balancing process, the aforementioned three methods minimize fairness and/or hitting costs by aggressively targeting locations with the lowest operational costs. {These approach inevitably incur a significantly higher switching cost. For instance, the average switching cost of \fairopt is still around 15 times that of the offline optimal benchmark.
In extreme cases, this ratio could be even higher, indicating considerable fluctuations in server provisioning—an undesirable strategy for data centers. It highlights the critical need to explicitly incorporate switching costs into the optimization framework}. \robd strategically balances between the minimizer of the hitting cost and the previous action, achieving the lowest switching cost while maintaining a reasonably low hitting cost. However, due to the non-separable nature of the long-term cost, \robd fails to minimize the fairness cost as a result of its algorithm design. 
By decomposing the long-term cost using auxiliary variables and dynamically optimizing these variables with mirror descent techniques, \ouralg achieves the best trade-off between hitting, switching, and fairness costs. This approach results in the lowest fairness and total costs among all the online algorithms.

To compare the impact of fairness cost, we present additional results by considering $u_3=1.75$ and $u_3=7$  on the performance of \ouralg, summarized 
in Tables~\ref{table:results_small_u} and~\ref{table:results_large_u}, respectively. 
Similar
as in Table~\ref{table:main_results}, we see that 
under three different learning rates $\eta$, \ouralg is very close to \opt in terms of the total cost regardless of the hyperparameters, which demonstrates the superior performance of \ouralg.
The other baselines (i.e., \fairopt, \hitmin, \robd, and \dmd)
all have substantially higher total costs. The other online baseline algorithms (\robd and \hitmin), which do not explicitly account for long-term fairness cost, consistently exhibit poor fairness performance across different values of the fairness cost parameter $u_3$. In contrast, \ouralg consistently maintains a relatively low fairness cost while simultaneously minimizing hitting cost and ensuring action smoothness. These results demonstrate that incorporating long-term fairness cost into our algorithm design is not only effective but also robust.

In summary, \ouralg effectively balances action smoothness while maintaining a low hitting cost and outperforms all other online baselines in fairness cost. These results validate the superiority of \ouralg’s explicit incorporation of long-term fairness in fostering socially responsible decision-making.

%% file: tex/appendix_proof_lb.tex
\section{Proof of the Lower Bounds}
\subsection{Proof of Theorem \ref{thm:regret_linear_bound}}\label{sec:regret_linear_bound_proof}
 
\begin{proof}

In this proof, we consider hitting, switching and long-term cost as following,
\begin{equation}
    cost(x_{1:T}) =  \frac{1}{T}\sum_{t=1}^T \frac{m_0}{2} \|x_t - c_t\|^2 +  \| \frac{1}{T} \sum_{t=1}^T A_t x_t\|_p , 
\end{equation}
where the horizon length $T$ is known in advance. For the hitting cost, we choose the constant value $m_0$ satisfying $m_0 >  1 $.
For an online policy, the context $c_t$ and $A_t$ are revealed sequentially,
and the future context can be chosen adversarially based on the actions from the online optimal policy. The offline policy has the complete information about the context.

Now we construct two selections for the online context sequence
\begin{equation}
\text{Option 1: }\left\{\begin{aligned}
c_{1:T}^* &= [\underbrace{1,1, \dots, 1}_{T/2}, \underbrace{0, 0, \dots, 0}_{T/2}] \\
A_{1:T}^* &=[\underbrace{1,1, \dots, 1}_{T/2}, \underbrace{0,0, \dots, 0}_{T/2}]
\end{aligned}\right. ,
\end{equation}

\begin{equation}
\text{Option 2: }\left\{\begin{aligned}
c_{1:T}^{\diamond } &= [\underbrace{1,1, \dots, 1}_{T}] \\
A_{1:T}^{\diamond } &=[\underbrace{1,1, \dots, 1}_{T/2}, \underbrace{-1,-1, \dots, -1}_{T/2}]
\end{aligned}\right. .
\end{equation}

Let $x_{1:T}^\dagger$ denote the action sequence generated by the online algorithm. For the first half of the horizon, $t \in [1, T/2]$, the two context sequences $(c_{1:T}^{*}, A_{1:T}^{*})$ and $(c_{1:T}^{\diamond}, A_{1:T}^{\diamond})$ are identical. The decision of which context sequence to choose is made at time $T/2 + 1$. 
Before revealing the context at this time, we examine the historical actions $x_{1:T/2}^\dagger$ of the expert policy.

For the online algorithm, we define the sum of action sequence as 
\begin{equation}
    \sum_{t=1}^{T/2} x_t^\dagger = \frac{T}{2} (S^\dagger + \epsilon(T)),
\end{equation}
where $S^\dagger$ is a constant value and $\epsilon(T)$ is a vanishing term satisfying $\lim_{T \rightarrow \infty}\epsilon(T) = 0$. We consider two cases about $S^\dagger$.

\textbf{Case 1}: $\lvert S^\dagger \rvert < \frac{2\,m_0-1}{2 \,m_0}$. 

In this case, we choose the second context trace $(c_{1:T}^{\diamond }, A_{1:T}^{\diamond })$. And the offline optimal action sequence can be obtained as 
\begin{equation}
    x_{1:T}^* = c_{1:T}^{\diamond }.
\end{equation}

And the total cost is calculated as 
\begin{equation}
    cost(x_{1:T}^*) = 0 +  \bigl\lvert \sum_{t=1}^T A_t^{\diamond }  c_t^{\diamond } \bigr\rvert = 0 .
\end{equation}

In this case, for any action sequence $x_{T/2+1:T}^\dagger$ chosen by the online algorithm, the cost of the expert algorithm is lower bounded by
\begin{equation}
\begin{aligned}
    cost(x_{1:T}^\dagger) &= \frac{1}{T}\sum_{t=1}^T \frac{m_0}{2} \|x_t^\dagger - c_t\|^2 +  \| \frac{1}{T} \sum_{t=1}^T A_t x_t\|_p \\
    &\geq \frac{1}{T}\sum_{t=1}^{T/2} \frac{m_0}{2} \|x_t^\dagger - 1\|^2\\ 
    &\geq \frac{m_0}{2\,T} \frac{T}{2} \biggl\| \frac{\sum_{t=1}^{T/2} x_t^\dagger 
 - T/2}{T/2}  \biggr\|^2  \\
 &= \frac{m_0}{4} \Bigl( S^\dagger - 1 +  \epsilon(T) \Bigr)^2  ,
\end{aligned}
\end{equation}
where the first inequality is due to the non-negativity of the hitting and long-term costs, the second inequality is based on RMS–AM inequality. Given $\epsilon(T)$ is a vanishing term, the regret is lower bounded by
\begin{equation}
    cost(x_{1:T}^\dagger) - cost(x_{1:T}^*) \gtrsim \frac{m_0}{4} \| S^\dagger - 1 \|^2 > \frac{m_0}{4} (\frac{1}{2\,m_0})^2 =  \frac{1}{16 m_0}   .
\end{equation}

\textbf{Case 2}: $\lvert S^\dagger \rvert \geq \frac{2\,m_0-1}{2\,m_0}$. 

In this case, we choose the first context trace $(c_{1:T}^{* }, A_{1:T}^{*})$. And one feasible action sequence is defined as 
\begin{equation}
    x_{1:T} = \frac{m_0-1}{m_0}\,c_{1:T}^{*} .
\end{equation}

We don't need to prove the optimality of this feasible action sequence, as any the offline optimal cost must be upper bounded by the cost of this feasible action sequence, which is
\begin{equation}
\begin{aligned}
    cost(x_{1:T}^*) &\leq \frac{1}{T}\sum_{t=1}^T \frac{m_0}{2} \|x_t - c_t\|^2 +  \| \frac{1}{T} \sum_{t=1}^T A_t x_t\|_p\\
    & = \frac{1}{T} \frac{m_0}{2} \frac{T}{2}\frac{1}{m_0^2} + \frac{1}{T} \frac{T}{2} \frac{m_0-1}{m_0} \\
    & = \frac{1}{4m_0} + \frac{m_0-1}{2m_0} = \frac{2m_0-1}{4m_0} .
\end{aligned}
\end{equation}

For the online algorithm, we have
\begin{equation}
\begin{aligned}
    cost(x_{1:T}^\dagger) &= \frac{1}{T}\sum_{t=1}^{T/2} \frac{m_0}{2} \|x_t^\dagger - c_t\|^2 +  \| \frac{1}{T} \sum_{t=1}^{T/2} A_t x_t\|_p \\
    &\geq \frac{1}{T}\sum_{t=1}^{T/2} \frac{m_0}{2} \|x_t^\dagger - 1\|^2 + 
 \frac{1}{2}\| S^\dagger + \epsilon(T)\|_p\\ 
    &\geq \frac{m_0}{2\,T} \frac{T}{2} \biggl\| \frac{\sum_{t=1}^{T/2} x_t^\dagger 
 - T/2}{T/2}  \biggr\|^2  + \frac{1}{2} \| S^\dagger + \epsilon(T)\|_p\\
 &= \frac{m_0}{4} \| {S^\dagger} - 1 +  \epsilon(T) \|^2 + \frac{1}{2}\| S^\dagger + \epsilon(T)\|_p\\
 &\gtrsim \frac{m_0}{4} ({S^\dagger} - 1)^2 + \frac{1}{2}\| S^\dagger\| .
\end{aligned}
\end{equation}

Then the regret of the online algorithm is lower bounded by
\begin{equation}
    cost(x_{1:T}^\dagger) - cost(x_{1:T}^*) \gtrsim \frac{m_0}{4} ({S^\dagger} - 1)^2 + \frac{1}{2} \| S^\dagger\| - \frac{2m_0-1}{4m_0} .
\end{equation}

Based on our condition on $S^\dagger$, if $S^\dagger  \leq -\frac{2\,m_0-1}{2\,m_0}$, it's obvious that 
\begin{equation}
    \frac{m_0}{4} ({S^\dagger} - 1)^2 + \frac{1}{2} \| S^\dagger\| - \frac{2m_0-1}{4m_0} \geq  \frac{m_0}{4} (1 + \frac{2\,m_0-1}{2\,m_0})^2 .
\end{equation}

On the other hand, if $S^\dagger > 0$, the minimizer of $\frac{m_0}{2} ({S^\dagger} - 1)^2 + \frac{1}{2} \| S^\dagger\| - \frac{2m_0-1}{4m_0} $ is achieved at the critical point $S^\dagger = 1 - \frac{1}{2 m_0}$.  
Therefore, $S^\dagger = 1 - \frac{1}{2\,m_0}$ is the minimizer for the bound as the bound monotonically increases as $S > 0$. So the lower bound in this case can be expressed as
\begin{equation}
    cost(x_{1:T}^\dagger) - cost(x_{1:T}^*) \gtrsim \frac{m_0}{4} (\frac{1}{2\,m_0})^2 + \frac{1}{2} \| 1 - \frac{1}{2\,m_0} \| - \frac{2m_0-1}{4m_0} = \frac{1}{16\,m_0} , 
\end{equation}

Given $m_0 >\frac{3}{2}$, 
the lower bounded for cost gap is obtained by combining the bounds for these two conditions for $S^\dagger$, which is
\begin{equation}
    cost(x_{1:T}^\dagger) - cost(x_{1:T}^*) \gtrsim \frac{1}{16m_0}.
\end{equation}

To sum up, compared with the offline optimal action, the regret of \textbf{any online algorithm} is lower bounded by $\Omega(1)$.

\end{proof}

\subsection{Proof of Theorem \ref{thm:cr_linear_bound}}
\begin{proof}
In this proof, we construct the hitting, switching and long-term cost as following,
\begin{equation}
    cost(x_{1:T}) =  \frac{1}{T}\sum_{t=1}^T \frac{m_0}{2} \|x_t - c_t\|^2 +  \| \frac{1}{T} \sum_{t=1}^T A_t x_t\|_p , 
\end{equation}
where the horizon length $T$ is known in advance and $m_0 > 1$. For an online policy, the context $c_t$ and $A_t$ are revealed sequentially,
and the future context can be chosen adversarially based on the actions from the online policy. The offline policy has the complete information about the context. 
Our goal is to derive a lower bound for the competitive ratio between online and offline policies under this information asymmetry. 

We construct two selections for the online context sequence
\begin{equation}
\text{Option 1: }\left\{\begin{aligned}
c_{1:T}^* &= [\underbrace{a,a, \dots, a}_{T/2}, \underbrace{0, 0, \dots, 0}_{T/2}] \\
A_{1:T}^* &=[\underbrace{1,1, \dots, 1}_{T/2}, 1, \underbrace{0,0, \dots, 0}_{T/2 - 1}]
\end{aligned}\right.  , 
\end{equation}

\begin{equation}
\text{Option 2: }\left\{\begin{aligned}
c_{1:T}^{\diamond } &= [\underbrace{a,a, \dots, a}_{T}] \\
A_{1:T}^{\diamond } &=[\underbrace{1,1, \dots, 1}_{T/2}, \underbrace{-1,-1, \dots, -1}_{T/2}]
\end{aligned}\right.   .
\end{equation}

Let $x_{1:T}^\dagger$ denote the action sequence generated by the online algorithm. For the first half of the horizon, $t \in [1, T/2]$, the two context sequences $(c_{1:T}^{*}, A_{1:T}^{*})$ and $(c_{1:T}^{\diamond}, A_{1:T}^{\diamond})$ are identical. The decision of which context sequence to choose is made at time $T/2 + 1$. 
Before revealing the context at this time, we examine the historical actions $x_{1:T/2}^\dagger$ of the online policy.

If there exists a time $t \in [1, T/2]$ such that $x_{t}^\dagger \neq a$, we select $(c_{1:T}^{\diamond}, A_{1:T}^{\diamond})$ as the context sequence. In this scenario, the offline optimal action is consistently $a$, resulting in a zero total cost. However, the expert's cost will be non-zero, leading to an infinite competitive ratio.

Conversely, if the online action sequence is $x_{1:T/2}^\dagger = [a,a,\cdots, a]$, we will choose $(c_{1:T}^{*}, A_{1:T}^{*})$ as the context sequence.  
The best an online algorithm can do is to only optimize the action in $x_{T/2 + 1}^\dagger$ and set all subsequent actions as zero. We define the objective for action at $x_{T/2 + 1}^\dagger$ as 
\begin{equation}
    \phi_1(x) = \frac{m_0}{2}(x)^2  + \bigl\| \frac{T}{2} a + x \bigr\|  . 
\end{equation}

In our current setting, $x$ is a scalar 
and the gradient of $g_1(x)$ is defined as
\begin{equation}
g_1'(x) =
\begin{cases}
  m_0 x + 1 &  x > -\frac{T\,a}{2} \\
  m_0 x - 1&  x < -\frac{T\,a}{2} \\
  \text{undefined}    & x = -1
\end{cases}    .
\end{equation}

In our following proof, we will choose the constant $a \geq \frac{2}{m_0\,T}$ (e.g. $a = \frac{2}{m_0\,T}$). 
To preserve generality, we will work with $a$ symbolically. 
Given that $m_0 > 0$ and $T > 2$, for all $x < -\frac{T\,a}{2}$, we can derive the upper bound of the gradient as follows, $g_1'(x) = m_0 x - 1 <  -m_0 \frac{T\,a}{2} - 1  < -1 $.
This implies that $g_1(x)$ is monotonically decreasing as $x$ approaches $-\frac{T\,a}{2}$ from the left. Since $\phi_1(x)$ is continuous, the minimizer cannot be located in the region $x < -\frac{T\,a}{2}$. 

On the other hand, as $x$ approaches $-\frac{T\,a}{2}$ from the right ($x > -\frac{T\,a}{2}$), the gradient $g_1'(x) = m_0 x + 1$ approaches $ - m_0 \frac{T\,a}{2} + 1 \leq -1  + 1 \leq  0$. Given $\phi_1(x)$ is continuous, this suggests that the minimizer of $\phi_1(x)$ must lie in the region $x \geq -\frac{T\,a}{2}$.
To find the critical point in this region, we set $g_1'(x) = 0$ for $x > -1$, which is $m_0 x + 1 = 0$. Solving for $x$, we obtain the potential minimizer:
\begin{equation}
    x_{T/2 + 1}^{\dagger,*} = - \frac{1}{m_0} . 
\end{equation}

And the minimum value of $g_1(x)$ becomes
\begin{equation}
\begin{aligned}
    g_1(x_{T/2 + 1}^{\dagger,*}) &= \frac{1}{T} \biggl[ \frac{m_0}{2}(x)^2  + \bigl\| \frac{T}{2} a + x \bigr\| \biggr]\\
    & = \frac{1}{T} \biggl[\frac{1}{2\,m_0} + \frac{T}{2} a - \frac{1}{m_0}\biggr]\\
    & = \frac{1}{2} a - \frac{1}{2m_0\,T},
\end{aligned}
\end{equation}
where the second equality is due to our choice of $a \geq \frac{2}{m_0\,T}$. 
The total cost of \textbf{any online algorithm} in this case is lower bounded by
\begin{equation}
    cost(x_{1:T}^\dagger) \geq  \frac{1}{2} a - \frac{1}{2m_0\,T}.
\end{equation}

Next, we aim to derive the upper bound for the offline policy with complete context sequence. As we only focus on the scale of the competitive ratio's lower bound, it's not necessary to explicitly solve the true offline optimal trace, which could be very messy. Instead, we choose the following action sequence to serve as a feasible action sequence $x_{1:T}'$, which is
\begin{equation}
    x_t' = \begin{cases}
         0 & t \in [1,\frac{T}{2}]\\
        -a & t = \frac{T}{2} + 1\\
        0 & t \in [\frac{T}{2}+1, T] , 
    \end{cases}
\end{equation}

Then the total cost of the true offline optimal cost is bounded by
\begin{equation}
\begin{aligned}
    cost(x_{1:T}^*)  \leq cost(x_{1:T}') &= \frac{1}{T} \biggl[ \frac{m_0}{2} (\frac{T}{2} + 1) a^2 + a \biggr]  .
\end{aligned}
\end{equation}

Here we choose $a = \frac{2}{m_0\,T}$. The lower bound for the cost of any online algorithm and the upper bound for the offline optimal cost is given by
\begin{equation}
\begin{aligned}
    cost(x_{1:T}^\dagger) &\geq  \frac{1}{2} a - \frac{1}{2m_0\,T}  = \frac{1}{2m_0\,T} ,\\
    cost(x_{1:T}^*)  &\leq \frac{1}{T} \biggl[\frac{m_0}{2} (\frac{T}{2} + 1)\,a^2 + a \biggr] = \frac{3}{m_0\,T^2} + \frac{2}{m_0\,T^3} . 
\end{aligned}
\end{equation}
The competitive ratio is lower bounded by
\begin{equation}
    CR =\frac{cost(x_{1:T}^\dagger)}{cost(x_{1:T}^*)} \geq \frac{\frac{1}{2m_0}}{\frac{3}{m_0\,T} + \frac{2}{m_0\,T^2}} = \frac{T}{6 + \frac{4}{T}} \gtrsim \frac{T}{6} .
\end{equation}
In other words, the competitive ratio of \textbf{any online algorithm} is lower bounded by $\Omega(T)$.
\end{proof}

%% file: tex/appendix_proof_acr.tex
\section{Proof of Asymptotic Competitive Ratio for \ouralg}

For the ease of presentation, we define the drift of Bregman divergence for the dual variable $\kappa$ at time $t$ as
\begin{equation}
    \Delta(t) = V_h(\kappa_1, \kappa_{t+1}) - V_h(\kappa_1, \kappa_{t})  .  
\end{equation}

To facilitate the proof of the asymptotic competitive ratio for \ouralg in different settings, we first present several technical lemmas.

\begin{lemma}\label{lemma:push_back}
    Suppose $\kappa_{t+1}$ is the solution to the following equation
    \begin{equation}
        \kappa_{t+1} = \arg\min_{\kappa \in \mathbb{R}^{N}} \langle d_t, \kappa \rangle + \frac{1}{\eta} V_h(\kappa, \kappa_t) .
    \end{equation}
    Then for any $\kappa' \in \mathbb{R}^{N} $, we have
    \begin{equation}
    \begin{aligned}
        &  \langle \kappa_{t+1}, d_t \rangle + \frac{1}{\eta}V_h(\kappa_{t+1}, \kappa_t)  \leq  \langle \kappa', d_t \rangle + \frac{1}{\eta}V_h(\kappa', \kappa_t) - \frac{1}{\eta}V_h(\kappa', \kappa_{t+1}) .
    \end{aligned}
    \end{equation}
\end{lemma}
\begin{proof}
    For simplicity, we define
    \begin{equation}
        e(\kappa) = \langle d_t, \kappa \rangle + \frac{1}{\eta} V_h(\kappa, \kappa_t) 
    \end{equation}
    The gradient of $e(\kappa)$ with respect to $\kappa$ is defined as
    \begin{equation}
        \nabla_{\kappa} e(\kappa) = d_t + \frac{1}{\eta} \bigl( \nabla h(\kappa) - \nabla h(\kappa_t) \bigr) . 
    \end{equation}
    Since $\kappa_{t+1}$ is the minimizer of $e(\kappa)$, then for any $\kappa' \in \mathbb{R}^{N} $ we must have
    \begin{equation}\label{eqn:gradient_inequality}
        \langle \kappa' - \kappa_{t+1}, d_t + \frac{1}{\eta} \bigl( \nabla h(\kappa_{t+1}) - \nabla h(\kappa_{t}) \bigr) \rangle \geq 0 . 
    \end{equation}
    We prove this by contradiction. Suppose there exist $\kappa'$, such that $\langle \kappa' - \kappa_{t+1}, \nabla_{\kappa} e(\kappa) \rangle < 0$,
    then we construct a function $y(\xi) = e( \kappa_{t+1} + \xi (\kappa' - \kappa_{t+1}) )$, where $\xi \in [0,1]$. The gradient of $y(\xi) $ is 
    \begin{equation}
    \begin{aligned}
        \nabla_{\xi} y(\xi) = \biggl\langle \kappa' - \kappa_{t+1}, d_t - \frac{1}{\eta} \Bigl( \nabla h(\kappa_{t}) - \nabla h \bigl(\kappa_{t+1} + \xi (\kappa' - \kappa_{t+1}) \bigr)
         \Bigr) \biggr\rangle .
    \end{aligned}
    \end{equation}
    Then $\nabla_{\xi} y(0) = \langle \kappa' - \kappa_{t+1}, d_t + \frac{1}{\eta} \bigl( \nabla h(\kappa_{t+1} ) - \nabla h(\kappa_{t}) \bigr)\rangle < 0 $. Therefore $\xi = 0$  is not the minimizer of 
 $y(\xi)$ or in other words,  $\kappa_{t+1}$ is not the minimizer of $e(\kappa)$, which is contradictory to our assumption. So the inequality in Eqn~\eqref{eqn:gradient_inequality} must hold for any $\kappa' \in \mathbb{R}^N$. 
By organizing the inequality, we have
    \begin{equation}
    \begin{aligned}
        \langle \kappa_{t+1}, d_t \rangle \leq & \langle \kappa', d_t \rangle + \frac{1}{\eta}\langle \kappa', \nabla h(\kappa_{t+1}) - \nabla h(\kappa_{t}) \rangle - \frac{1}{\eta}\langle \kappa_{t+1}, \nabla h(\kappa_{t+1}) - \nabla h(\kappa_{t}) \rangle \\
        = & \langle \kappa', d_t \rangle + \frac{1}{\eta}\langle \kappa' - \kappa_{t+1}, \nabla h(\kappa_{t+1}) - \nabla h(\kappa_{t}) \rangle  .
    \end{aligned}
    \end{equation}
    According to the definition of Bregman Divergence, we have
    \begin{equation}
    \begin{aligned}
        & \langle \kappa' - \kappa_{t+1}, \nabla h(\kappa_{t+1}) - \nabla h(\kappa_{t}) \rangle \\
        = & \langle \kappa' - \kappa_{t+1}, \nabla h(\kappa_{t+1})  \rangle - \langle \kappa' - \kappa_{t+1}, \nabla h(\kappa_{t}) \rangle \\
        = & \langle \kappa' - \kappa_{t+1}, \nabla h(\kappa_{t+1})  \rangle - \langle \kappa' - \kappa_{t}, \nabla h(\kappa_{t}) \rangle  + \langle  \kappa_{t+1} - \kappa_{t}, \nabla h(\kappa_{t}) \rangle \\
        = & - h(\kappa') + h(\kappa_{t+1}) + \langle \kappa'  - \kappa_{t+1}, \nabla h(\kappa_{t+1})  \rangle - h(\kappa_t) + h(\kappa') - \langle \kappa' - \kappa_{t}, \nabla h(\kappa_{t}) \rangle  \\
        & - h(\kappa_{t+1}) + h(\kappa_t) + \langle  \kappa_{t+1} - \kappa_{t}, \nabla h(\kappa_{t}) \rangle\\
        = & -V_{h}(\kappa', \kappa_{t+1}) + V_h(\kappa', \kappa_{t}) - V_h(\kappa_{t+1}, \kappa_{t}) .
    \end{aligned}
    \end{equation}
    Then we finish the proof. 
\end{proof}

\begin{lemma}\label{lemma:queue_length_general_convex}
    If the reference function $h(\cdot)$ is $l$-strongly convex and $\kappa_{1}$ is the initial dual variable, then the dual variable $\kappa_t$ is bounded by
    \begin{equation}
        V_h(\kappa_1, \kappa_{T+1}) \leq   T( \eta^2 \frac{Z^2}{2l} + \eta L Z) - \kappa_1^\top \cdot (\sum_{t=1}^T A_t  x_t  - z_t) , 
    \end{equation}
    where the constant $Z = \sup_{z_t, x_t, A_t}\|z_t - A_t x_t\|$. 
\end{lemma}
\begin{proof}
    According to the Line 7 of Algorithm \ref{alg:algorithm_main} and Lemma \ref{lemma:push_back}, for any $\kappa' \in \mathbb{R}_{+}^N$ we have
    \begin{equation}\label{eqn:update_rule_ineq}
    \begin{aligned}
        & \langle \kappa_{t+1}, d_t \rangle + \frac{1}{\eta}V_h(\kappa_{t+1}, \kappa_t)
        \leq \langle \kappa', d_t \rangle + \frac{1}{\eta}V_h(\kappa', \kappa_t) - \frac{1}{\eta}V_h(\kappa', \kappa_{t+1})  .
    \end{aligned}
    \end{equation} 
    By setting $\kappa' = \kappa_1$ and add $\langle \kappa_{t}, d_t \rangle$ to both sides of Eqn~\eqref{eqn:update_rule_ineq}, we have
    \begin{equation}\label{eqn:update_rule_ineq_2}
    \begin{aligned}
        \langle \kappa_{t} - \kappa_1, d_t \rangle  \leq & \left( \langle \kappa_t - \kappa_{t+1}, d_t \rangle 
 - \frac{1}{\eta}V_h(\kappa_{t+1}, \kappa_t) \right) \\
 & + \frac{1}{\eta} V_h(\kappa_1, \kappa_t) - \frac{1}{\eta}V_h(\kappa_1, \kappa_{t+1}) . 
    \end{aligned}
    \end{equation}
Since the reference function $h(a)$ is $l$-strongly convex, then we have
\begin{equation}
    h(a) \geq h(b) + (a-b) \nabla h(b) + \frac{l}{2}\|a - b\|^2 .
\end{equation}
In other words, 
\begin{equation}
    V_h(\kappa_{t+1}, \kappa_t) \geq \frac{l}{2}\|\kappa_{t+1} - \kappa_t\|^2 . 
\end{equation}
Therefore, we have
\begin{equation}
\begin{aligned}
    & \frac{1}{\eta}V_h(\kappa_{t+1}, \kappa_t) - \langle \kappa_t - \kappa_{t+1}, d_t \rangle \\
    = & \langle \kappa_{t+1} - \kappa_t , d_t \rangle + \frac{1}{\eta}V_h(\kappa_{t+1}, \kappa_t) \\
    \geq &- \| \kappa_{t+1} - \kappa_t \| \| d_t \| + \frac{l}{2\eta} \|\kappa_{t+1} - \kappa_t\|^2 \\
    \geq &- \frac{l}{2\eta}\| \kappa_{t+1} - \kappa_t \|^2 - \frac{\eta}{2l}\| d_t \|^2  + \frac{l}{2\eta} \|\kappa_{t+1} - \kappa_t\|^2\\
    = & - \frac{\eta}{2l}\| d_t \|^2 \geq - \frac{\eta}{2l} Z^2 . 
\end{aligned}
\end{equation}
The first inequality holds by the definition of inner product and the $l$-strongly convexity assumption of reference function $h(\cdot)$, the second inequality is derived with the AM-GM inequality, and the last inequality is obtained with the assumption $\|d_t\| \leq Z$. According to the definition of $\Delta(t)$, Eqn~\eqref{eqn:update_rule_ineq_2} becomes
\begin{equation}\label{eqn:singl_step_dual_gap}
    \langle \kappa_{t} , z_t - A_t x_t \rangle + \frac{\Delta(t)}{\eta}  \leq \frac{\eta}{2l} Z^2 + \langle \kappa_1, z_t - A_t x_t  \rangle . 
\end{equation}
According to Line 4 and Line 5 of algorithm \ref{alg:algorithm_main}, then for any $x_t' \in \mathcal{X}_t$ and $z_t' \in \mathcal{Z}$, we have 
\begin{equation}
\begin{aligned}
    & f_t(x_t) + \lambda_1 d(x_t, x_{t-1}) + \frac{\lambda_2}{2} \| x_t - v_t \|^2  +  g(z_t) + \kappa_t^\top \cdot ( A_t  x_t  - z_t )\\
    \leq & f_t(x_t') + \lambda_1 d(x_t', x_{t-1}) + \frac{\lambda_2}{2} \| x_t' - v_t \|^2  +  g(z_t') +  \kappa_t^\top \cdot ( A_t  x_t'  - z_t' )
\end{aligned}
\end{equation}
By adding $\biggl(\frac{\Delta(t)}{\eta} - \kappa_t^\top \cdot ( A_t  x_t  - z_t ) \biggr)$ to both sides of the inequality, we have
\begin{equation}
\begin{aligned}
    & \frac{\Delta(t)}{\eta} + f_t(x_t) + \lambda_1 d(x_t, x_{t-1}) + \frac{\lambda_2}{2} \| x_t - v_t \|^2 + g(z_t)   \\
    \leq & \biggl(\frac{\Delta(t)}{\eta} - \kappa_t^\top \cdot ( A_t  x_t  - z_t ) \biggr) + f_t(x_t') + \lambda_1 d(x_t', x_{t-1}) + \frac{\lambda_2}{2} \| x_t' - v_t \|^2  +  g(z_t') + \kappa_t^\top \cdot ( A_t  x_t'  - z_t' ) \\
    \leq & \left( \frac{\eta}{2l}Z^2 -  \kappa_1^\top \cdot ( A_t  x_t - z_t ) \right)+ f_t(x_t') + \lambda_1 d(x_t', x_{t-1}) + \frac{\lambda_2}{2} \| x_t' - v_t \|^2  +  g(z_t') + \kappa_t^\top \cdot ( A_t  x_t'  - z_t' ) .
\end{aligned}
\end{equation}
Now we choose $x_t' = x_t$ and $z_t' = A_t x_t'$, then we have
\begin{equation}
\begin{aligned}
     \frac{\Delta(t)}{\eta} \leq & \left( \frac{\eta}{2l}Z^2 -  \kappa_1^\top \cdot ( A_t  x_t - z_t ) \right) + g(z_t') - g(z_t) \leq  \frac{\eta}{2l}Z^2 + LZ - \kappa_1^\top \cdot (  A_t  x_t - z_t ) .
\end{aligned}
\end{equation}
By summing up the difference over the episode $T$, we have
\begin{equation}
\begin{aligned}
    & V_h(\kappa_{1}, \kappa_{T+1}) - V_h(\kappa_1, \kappa_1) 
    \leq \eta T(\frac{\eta}{2l}Z^2 + LZ) - \eta \kappa_1^\top \cdot (\sum_{t=1}^T A_t  x_t  - z_t) 
\end{aligned}
\end{equation}
\end{proof}

\subsection{Proof of Theorem \ref{thm:dmd_algorithm_no_switch}}

In the problem setting without switching cost, we define the optimization objective as 
\begin{equation}
    G({x}_{1:T}) = \frac{1}{T}\sum_{t=1}^T \biggl[ f_t({x}_t)\biggr] +   g(\frac{1}{T}\sum_{t=1}^{T} A_t {x}_t).
\end{equation}

\begin{proof}
Since $x_t$, $z_t$ is the optimal solution for Eqn~\eqref{eqn:single_opt_1}, so we have the following equation for any $x_t' \in \mathcal{X}, z_t' \in \mathcal{Z}$:
 \begin{equation}\label{eqn:single_step_minimizer_no_switch}
    \begin{aligned}
         f_t(x_t) + \kappa_t \cdot A_t  x_t 
        \leq f_t(x_t') + \kappa_t \cdot A_t  x_t' , 
    \end{aligned}
    \end{equation}
 \begin{equation}\label{eqn:single_step_aux_no_switch}
        g(z_t) - \kappa_t z_t \leq g(z_t') - \kappa_t z_t' . 
    \end{equation}
By combining the inequalities in Eqn~\eqref{eqn:singl_step_dual_gap},~\eqref{eqn:single_step_minimizer_no_switch} and~\eqref{eqn:single_step_aux_no_switch}, we have
    \begin{equation}
        \begin{aligned}\label{eqn:single_inequality_1_no_switch}
            \frac{\Delta(t)}{\eta} + f_t(x_t)  + g(z_t)
            \leq  \frac{\eta}{2l} Z^2 + f_t(x_t')  + g(z_t') + \kappa_t^\top \cdot ( A_t  x_t'  - z_t' ) - \kappa_1^\top \cdot  ( A_t  x_t  - z_t ) .
        \end{aligned}
    \end{equation}

For the last term of Eqn~\eqref{eqn:single_inequality_1_no_switch}, since $\kappa_{t+1}$ is the minimizer of $e(\kappa)$, then for any $\kappa' \in \mathbb{R}^{N} $ we must have
\begin{equation}
    \langle \kappa' - \kappa_{t+1}, d_t + \frac{1}{\eta} \bigl( \nabla h(\kappa_{t+1}) - \nabla h(\kappa_{t}) \bigr) \rangle \geq 0 . 
\end{equation}
This condition is satisfied only if the gradient is zero, which is
\begin{equation}
    d_t + \frac{1}{\eta} \bigl( \nabla h(\kappa_{t+1}) - \nabla h(\kappa_{t}) \bigr) = 0 .
\end{equation}
Since the reference function is $l$-strongly convex, we can bound the dual update by the gradient distance in the following way
\begin{equation}
    \| \kappa_{t+1} - \kappa_{t}\| \leq \frac{\eta}{l} \| \bigl( \nabla h(\kappa_{t+1}) - \nabla h(\kappa_{t}) \bigr)\| \leq \frac{\eta\| d_t \|}{l} . 
\end{equation}
Thus, we can have the following inequality for the last term of Eqn~\eqref{eqn:single_inequality_1_no_switch} by summing it up over the frame $R$, which is
\begin{equation}
\begin{aligned}
    \langle \kappa_{t+k} - \kappa_{t}, A_{t+k}  x_{t+k}'  - z_{t+k}' \rangle 
    \leq& \Bigl\| \sum_{j=t}^{t+k-1} \bigl( \kappa_{j+1} -\kappa_{j}\bigr) \Bigr\| \cdot \bigl\| A_{t+k}  x_{t+k}'  - z_{t+k}' \bigr\| \\
    \leq & \frac{\eta}{l} \cdot \Bigl( \sum_{j=t}^{t+k-1} \bigl\|   A_j x_j - z_j  \bigr\| \Bigr) \cdot Z \leq (k-1)\cdot \frac{\eta Z^2}{l}.
\end{aligned}
\end{equation}
Then within the frame of size $R$, we have
\begin{equation}\label{eqn:dual_bound_no_switch}
\begin{aligned}
    & \left[\sum_{t=kR+1}^{kR+R} \kappa_t \cdot ( A_t  x_t'  - z_t' )\right]  
    \leq \kappa_{kR+1} \left[  \sum_{t=kR+1}^{kR+R} ( A_t  x_t'  - z_t' ) \right] + \frac{R(R-1)}{2 l}\eta Z^2 .
\end{aligned}
\end{equation}
Then by substituting Eqn~\eqref{eqn:dual_bound_no_switch} to Eqn~\eqref{eqn:single_inequality_1_no_switch} and summing up over the whole episode $T$, we have:
    \begin{equation}\label{eqn:total_cost_ineq_no_switch}
        \begin{aligned}
            &\sum_{t=1}^T \frac{\Delta(t)}{\eta} +  \sum_{t=1}^T \biggl[ f_t(x_t) + g(z_t)\biggr] 
            = \frac{V_h(\kappa_1, \kappa_{T+1})}{\eta} + \sum_{t=1}^T \biggl[ f_t(x_t) + g(z_t)\biggr] \\
            &\leq  \sum_{t=1}^T \biggl[ f_t(x_t') + g(z_t') \biggr]  +\frac{\eta}{2l} Z^2 T R   - \kappa_1^\top \cdot \left[ \sum_{t=1}^T A_t x_t - z_t \right]
            +\left[ \sum_{k=1}^{K} \kappa_{(k-1)R+1} \sum_{t=(k-1)R+1}^{kR}  ( A_t  x_t'  - z_t' ) \right] 
        \end{aligned}
    \end{equation}

Now suppose $x_{1:T}^*$ is the $(R, \delta)$-optimal action sequence, then within the $k$-th frame we manually construct $z_t^*$ as 
\begin{equation}
    z_t^* = \frac{1}{R} \sum_{i=kR+1}^{kR+R} A_i  x_i^*, \; \forall t \in [kR+1, kR+R], 1\leq k \leq K .
\end{equation}

Based on the $L$-Lipschitz assumption on function $g(\cdot)$, we have:
\begin{equation}
\begin{aligned}
    \sum_{t=1}^T g(z_t^*) = & \sum_{k=1}^{K} R \cdot g(\frac{1}{R} \sum_{t=(k-1)R+1}^{kR} A_t  x_t^*)\\
    \leq & R \sum_{k=1}^{K}  g(\frac{1}{T}\sum_{t=1}^{T} A_t x_t^*)   +  L \sum_{k=1}^{K} \| \frac{1}{R}\sum_{t=(k-1)R+1}^{kR} A_t  x_t^*  - \frac{1}{T}\sum_{t=1}^{T} A_t x_t^* \| \\
    \leq & \sum_{k=1}^{K} \Bigl[ R \cdot g(\frac{1}{T}\sum_{t=1}^{T} A_t x_t^*) \Bigr] + L\delta \\
    = & T g(\frac{1}{T}\sum_{t=1}^{T} A_t x_t^*)  + L \delta  .
\end{aligned}
\end{equation}
Substituting it back to Eqn~\eqref{eqn:total_cost_ineq_no_switch}, we have
\begin{equation}\label{eqn:objective_difference_1_no_switch}
    \begin{aligned}
        \sum_{t=1}^T \biggl[ f_t(x_t) +  g(z_t)  \biggr]
        \leq & \sum_{t=1}^T \biggl[f_t(x_t^*)\biggr]  +  T \cdot g(\frac{1}{T}\sum_{t=1}^{T} A_t x_t^*)  +
         \frac{\eta}{2l} Z^2 T R\\
         &+ L\delta - \frac{V_h(\kappa_{1}, \kappa_{T+1}) }{\eta} - \kappa_1^\top \cdot  \left[ \sum_{t=1}^T A_t x_t - z_t \right] . 
    \end{aligned}
\end{equation}
where the Bregman divergence $V_h(\kappa_{1}, \kappa_{T+1})$ is non-negative since the reference function is convex, which can be eliminated later. The last step is to upper bound $g(\frac{1}{T}\sum_{t=1}^T A_t x_t)$ with $\frac{1}{T}\sum_{t=1}^T g(z_t)$, which is 
\begin{equation}\label{eqn:fairness_bound_no_switch}
\begin{aligned}
    &T \cdot g(\frac{1}{T}\sum_{t=1}^T A_t x_t) - \sum_{t=1}^T g(z_t)\\
    \leq & T L \| \frac{1}{T} \sum_{t=1}^T A_t x_t - \frac{1}{T}\sum_{t=1}^T z_t\|  + T \biggl[ g(\frac{1}{T}\sum_{t=1}^T z_t) - \frac{1}{T} \sum_{t=1}^T g(z_t) \biggr] \\
    \leq & L \| \sum_{t=1}^T A_t x_t - \sum_{t=1}^T z_t\| . 
\end{aligned}
\end{equation}
Since $\kappa_{t+1}$ is the minimizer of $e(\kappa)$, we have
\begin{equation}
    d_t + \frac{1}{\eta} \bigl( \nabla h(\kappa_{t+1}) - \nabla h(\kappa_{t}) \bigr) = 0 .
\end{equation}
    
In other words
\begin{equation}
    \Bigl\| \sum_{t=1}^T (- d_t) \Bigr\| = \Bigl\| \sum_{t=1}^T \bigl(A_t x_t - z_t \bigr) \Bigr\| =  \frac{\| \nabla h(\kappa_{T+1}) - \nabla h(\kappa_{1}) \| }{\eta}   .
\end{equation}
Since the reference function $h(\cdot)$ is $l$-strongly convex and $\beta_2$-smooth, we have
\begin{equation}
\begin{aligned}
    &\frac{\| \nabla h(\kappa_{T+1}) - \nabla h(\kappa_{1}) \| }{\eta} \leq \frac{\beta_2 \| \kappa_{T+1} -  \kappa_{1} \| }{\eta}  
    \leq  \frac{\beta_2}{\eta} \sqrt{\frac{2}{l} V_h(\kappa_1, \kappa_{T+1})} .
\end{aligned}
\end{equation}
By substituting the inequality to Eqn~\eqref{eqn:fairness_bound_no_switch}, we have
\begin{equation}
\begin{aligned}
    T \cdot g(\frac{1}{T}\sum_{t=1}^T A_t x_t) - \sum_{t=1}^T g(z_t)  \leq   \frac{\beta_2 L}{\eta} \sqrt{\frac{2}{l} V_h(\kappa_1, \kappa_{T+1})} . 
\end{aligned}
\end{equation}
By substituting Lemma \ref{lemma:queue_length_general_convex} into the inequality, we have
\begin{equation}
\begin{aligned}
     & T \cdot g(\frac{1}{T}\sum_{t=1}^T A_t x_t) - \sum_{t=1}^T g(z_t)   
     \leq {\beta_2 L} \sqrt{ T(  \frac{Z^2}{l^2} +  \frac{1}{\eta} \frac{2 L Z}{l} )-  \frac{2}{\eta^2 l} \kappa_1^\top \cdot (\sum_{t=1}^T A_t  x_t  - z_t)} . 
\end{aligned}
\end{equation}
According to our assumption, we have $\Bigl| \kappa_1^\top \cdot (\sum_{t=1}^T A_t  x_t  - z_t) \Bigr| \leq T Z \cdot \| \kappa_1 \|$.
By substituting the above inequality into Eqn~\eqref{eqn:objective_difference_1_no_switch}, we have
\begin{equation}
    \begin{aligned}
        \sum_{t=1}^T \biggl[ f_t(x_t)   \biggr]  + T \cdot g(\frac{1}{T}\sum_{t=1}^T A_t x_t)
        \leq &   \sum_{t=1}^T \biggl[ f_t(x_t^*)  \biggr]  + T \cdot g(\frac{1}{T}\sum_{t=1}^{T} A_t x_t^*)    + \frac{\eta}{2l} Z^2 T R  \\
        & + TZ\|\kappa_1\| + L\delta + {\beta_2 L} \sqrt{ T \Bigl( \frac{Z^2}{l^2} +  \frac{1}{\eta} \frac{2 L Z}{l}  + \frac{2}{\eta^2 l} Z \| \kappa_1 \| \Bigr) }  . 
    \end{aligned}
\end{equation}
By dividing both sides by $T$, we finish the proof.
\end{proof}

\subsection{Proof of Theorem \ref{thm:dmd_algorithm_switch}}

For the ease of presentation, we define the optimization objective $G({x}_{1:T}, \tilde{x}_{0:T-1})$ as
\begin{equation}
\begin{aligned}
    G({x}_{1:T}, \tilde{x}_{0:T-1}) = \frac{1}{T}\sum_{t=1}^T \biggl[ f_t({x}_t) + \lambda_1 d({x}_t, \tilde{x}_{t-1})  + \frac{\lambda_2}{2} \| {x}_t - v_t \|^2 \biggr] +   g(\frac{1}{T}\sum_{t=1}^{T} A_t {x}_t) . 
\end{aligned}
\end{equation}
To facilitate the proof of the asymptotic competitive ratio for \ouralg when the hitting cost is $m$-strongly convex, we first present several technical lemmas. 
\begin{lemma}\label{lemma:total_cost}
     Suppose the switching cost is defined as as $d(x_t, x_{t-1}) = \frac{ \beta_1}{2}\| x_t -  x_{t-1}\|^2$ and the action sequence $\{({x}_t, {z}_t) | \forall t\in [1,T] \}$ is the solution using Algorithm~\ref{alg:algorithm_main}, then for any $x(t)'\in \mathcal{X}$, $z(t)' \in \mathcal{Z}$, we have
    \begin{equation}\label{eqn:total_cost_ineq}
        \begin{aligned}
            &\sum_{t=1}^T \frac{\Delta(t)}{\eta} +  \sum_{t=1}^T \biggl[ f_t(x_t) + \lambda_1 d(x_t, x_{t-1}) + g(z_t) \biggr]  +  \sum_{t=1}^T \frac{\lambda_2}{2} \| x_t - v_t \|^2  \\
            \leq &   \sum_{t=1}^T \biggl[ f_t(x_t') + \lambda_1 d(x_t', x_{t-1}) + \frac{\lambda_2}{2} \| x_t' - v_t \|^2  +  g(z_t') \biggr]  + \left[ \sum_{k=1}^{K} \kappa_{(k-1)R+1} \sum_{t=(k-1)R+1}^{kR}  ( A_t  x_t'  - z_t' ) \right] \\
            &  - \frac{m+\lambda_1  \beta_1 + \lambda_2}{2}\sum_{t=1}^T \| x_t' - x_t\|^2  + \frac{\eta}{2l} Z^2 T R   - \kappa_1^\top \cdot \left[ \sum_{t=1}^T A_t x_t - z_t \right] . 
        \end{aligned}
    \end{equation}
\end{lemma}
\begin{proof}
    Since the function $f_t(x)$ is $m$-strongly convex and switching cost $d(x, x_{t-1})$is also  {$\beta_1$}-strongly convex with respect to $x$ by definition, then we have
    \begin{equation}\label{eqn:single_step_hit_switch}
    \begin{aligned}
        & f_t(x_t) + \lambda_1 d(x_t, x_{t-1}) + \frac{\lambda_2}{2} \| x_t - v_t \|^2  + \kappa_t \cdot A_t  x_t   + \frac{m+\lambda_1 \beta_1+ \lambda_2}{2} \|x_t' - x_t \|^2\\
        \leq &f_t(x_t') + \lambda_1 d(x_t', x_{t-1}) + \frac{\lambda_2}{2} \| x_t' - v_t \|^2  + \kappa_t \cdot A_t  x_t' . 
    \end{aligned}
    \end{equation}
    where $x_t$ is the optimal solution for Eqn~\eqref{eqn:single_opt_1}. Then similarly, for $z_t$, we have
    \begin{equation}\label{eqn:single_step_aux}
        g(z_t) - \kappa_t z_t \leq g(z_t') - \kappa_t z_t' . 
    \end{equation}
    By combining the inequalities in Eqn~\eqref{eqn:singl_step_dual_gap},~\eqref{eqn:single_step_hit_switch} and~\eqref{eqn:single_step_aux}, we have
    \begin{equation}
        \begin{aligned}\label{eqn:single_inequality_1}
            &\frac{\Delta(t)}{\eta} + f_t(x_t) + \lambda_1 d(x_t, x_{t-1}) + \frac{\lambda_2}{2} \| x_t - v_t \|^2  + g(z_t) + \frac{m+\lambda_1 \beta_1 + 
        \lambda_2}{2} \|x_t' - x_t \|^2\\
            \leq & \frac{\eta}{2l} Z^2 + f_t(x_t') + \lambda_1 d(x_t', x_{t-1}) + \frac{\lambda_2}{2} \| x_t' - v_t \|^2   + g(z_t') + \kappa_t^\top \cdot ( A_t  x_t'  - z_t' ) - \kappa_1^\top \cdot  ( A_t  x_t  - z_t ). 
        \end{aligned}
    \end{equation}
    For the last term of Eqn~\eqref{eqn:single_inequality_1}, since $\kappa_{t+1}$ is the minimizer of $e(\kappa)$, then for any $\kappa' \in \mathbb{R}^{N} $ we must have
    \begin{equation}
        \langle \kappa' - \kappa_{t+1}, d_t + \frac{1}{\eta} \bigl( \nabla h(\kappa_{t+1}) - \nabla h(\kappa_{t}) \bigr) \rangle \geq 0 . 
    \end{equation}
    This condition is satisfied only if the gradient is zero, which is
    \begin{equation}
        d_t + \frac{1}{\eta} \bigl( \nabla h(\kappa_{t+1}) - \nabla h(\kappa_{t}) \bigr) = 0 .
    \end{equation}
    Since the reference function is $l$-strongly convex, we can bound the dual update by the gradient distance in the following way
    \begin{equation}
        \| \kappa_{t+1} - \kappa_{t}\| \leq \frac{\eta}{l} \| \bigl( \nabla h(\kappa_{t+1}) - \nabla h(\kappa_{t}) \bigr)\| \leq \frac{\eta\| d_t \|}{l} . 
    \end{equation}
    Thus, we can have the following inequality for the last term of Eqn~\eqref{eqn:single_inequality_1} by summing it up over the frame $R$, which is
    \begin{equation}
    \begin{aligned}
        \langle \kappa_{t+k} - \kappa_{t}, A_{t+k}  x_{t+k}'  - z_{t+k}' \rangle 
        \leq& \Bigl\| \sum_{j=t}^{t+k-1} \bigl( \kappa_{j+1} -\kappa_{j}\bigr) \Bigr\| \cdot \bigl\| A_{t+k}  x_{t+k}'  - z_{t+k}' \bigr\| \\
        \leq & \frac{\eta}{l} \cdot \Bigl( \sum_{j=t}^{t+k-1} \bigl\|   A_j x_j - z_j  \bigr\| \Bigr) \cdot Z \leq (k-1)\cdot \frac{\eta Z^2}{l}.
    \end{aligned}
    \end{equation}
    Then within the frame of size $R$, we have
    \begin{equation}\label{eqn:dual_bound}
    \begin{aligned}
        & \left[\sum_{t=kR+1}^{kR+R} \kappa_t \cdot ( A_t  x_t'  - z_t' )\right]  
        \leq \kappa_{kR+1} \left[  \sum_{t=kR+1}^{kR+R} ( A_t  x_t'  - z_t' ) \right] + \frac{R(R-1)}{2 l}\eta Z^2 .
    \end{aligned}
    \end{equation}
    Then we finish the proof by substituting Eqn~\eqref{eqn:dual_bound} to Eqn~\eqref{eqn:single_inequality_1} and summing up over the whole episode $T$. 
 
\end{proof}

\begin{lemma}\label{lemma:intermediate_cost}
    Given any context sequence $\gamma = \{ (f_t(\cdot), A_t) | \forall t\in [1,T] \}$, $x_{1:T}$ and $x_{1:T}^*$  are the action sequence of \ouralg and the $(R, \delta)$-optimal benchmark, respectively. Suppose the switching cost $d(x_t, x_{t-1}) = \frac{ \beta_1}{2}\| x_t -  x_{t-1}\|^2$, then the optimization objective of \ouralg is bounded by
    \begin{equation}\label{eqn:cost_gap_1}
    \begin{aligned}
        G({x}_{1:T}, {x}_{0:T-1})  
        \leq & G({x}_{1:T}^*, {x}_{0:T-1}) 
          - \frac{m+\lambda_1  \beta_1 + \lambda_2}{2} \frac{1}{T} \sum_{t=1}^T \| x_t^* - x_t\|^2  \\
         & + \frac{\eta}{2l} Z^2 R   + Z \| \kappa_1\| +  \frac{L\delta}{T}  + \beta_2 L \sqrt{\frac{1}{T}(\frac{Z^2}{l^2} + 2\frac{L Z}{\eta l} + \frac{2}{\eta^2 l} Z \|\kappa_1 \|)}  . 
    \end{aligned}
\end{equation}
\end{lemma}
\begin{proof}
    With Lemma \ref{lemma:queue_length_general_convex} and Lemma \ref{lemma:total_cost}, we are ready to prove the Lemma \ref{lemma:intermediate_cost}. Now suppose $x_{1:T}^*$ is the $(R, \delta)$-optimal action sequence, then within the $k$-th frame we manually construct $z_t^*$ as 
\begin{equation}
    z_t^* = \frac{1}{R} \sum_{i=kR+1}^{kR+R} A_i  x_i^*, \; \forall t \in [kR+1, kR+R], 1\leq k \leq K
\end{equation}
Then according to Lemma~\ref{lemma:total_cost}, we have 

\begin{equation}\label{eqn:optimal_bound_1}
    \begin{aligned}
        &\frac{V_h( \kappa_{1}, \kappa_{T+1})  }{\eta} +  \sum_{t=1}^T \biggl[ f_t(x_t) + \lambda_1 d(x_t, x_{t-1}) +  g(z_t)  \biggr] + \sum_{t=1}^T \frac{\lambda_2}{2} \| x_t - v_t \|^2  \\
        \leq &   \sum_{t=1}^T \biggl[ f_t(x_t^*) + \lambda_1 d(x_t^*, x_{t-1}) + \frac{\lambda_2}{2} \| x_t^* - v_t \|^2  +  g(z_t^*) \biggr]  \\
        & - \frac{m+\lambda_1 \beta_1 + \lambda_2}{2}\sum_{t=1}^T \| x_t^* - x_t\|^2 + \frac{\eta}{2l} Z^2 T R  - \kappa_1^\top \cdot \left[ \sum_{t=1}^T A_t x_t - z_t \right] . 
    \end{aligned}
\end{equation}
Besides, based on the $L$-Lipschitz assumption on function $g(\cdot)$, we have

\begin{equation}
\begin{aligned}
    \sum_{t=1}^T g(z_t^*) = & \sum_{k=1}^{K} R \cdot g(\frac{1}{R} \sum_{t=(k-1)R+1}^{kR} A_t  x_t^*)\\
    \leq & R \sum_{k=1}^{K}  g(\frac{1}{T}\sum_{t=1}^{T} A_t x_t^*)   +  L \sum_{k=1}^{K} \| \frac{1}{R}\sum_{t=(k-1)R+1}^{kR} A_t  x_t^*  - \frac{1}{T}\sum_{t=1}^{T} A_t x_t^* \| \\
    \leq & \sum_{k=1}^{K} \Bigl[ R \cdot g(\frac{1}{T}\sum_{t=1}^{T} A_t x_t^*) \Bigr] + L\delta \\
    = & T g(\frac{1}{T}\sum_{t=1}^{T} A_t x_t^*)  + L \delta . 
\end{aligned}
\end{equation}
Substituting it back to Eqn~\eqref{eqn:optimal_bound_1}, we have
\begin{equation}\label{eqn:objective_difference_1}
    \begin{aligned}
        & \sum_{t=1}^T \biggl[ f_t(x_t) + \lambda_1 d(x_t, x_{t-1}) + \frac{\lambda_2}{2} \| x_t - v_t \|^2  +  g(z_t)  \biggr]\\
        \leq &   \sum_{t=1}^T \biggl[ f_t(x_t^*) + \lambda_1 d(x_t^*, x_{t-1}) + \frac{\lambda_2}{2} \| x_t^* - v_t \|^2 \biggr] +  T \cdot g(\frac{1}{T}\sum_{t=1}^{T} A_t x_t^*)   \\
        &  - \frac{m+\lambda_1  \beta_1+ \lambda_2}{2}\sum_{t=1}^T \| x_t^* - x_t\|^2  + \frac{\eta}{2l} Z^2 T R+ L\delta - \frac{V_h(\kappa_{1}, \kappa_{T+1}) }{\eta} - \kappa_1^\top \cdot  \left[ \sum_{t=1}^T A_t x_t - z_t \right] . 
    \end{aligned}
\end{equation}
 
where the Bregman divergence $V_h(\kappa_{1}, \kappa_{T+1})$ is non-negative since the reference function is convex, which can be eliminated later. The last step is to upper bound $g(\frac{1}{T}\sum_{t=1}^T A_t x_t)$ with $\frac{1}{T}\sum_{t=1}^T g(z_t)$, which is 
\begin{equation}\label{eqn:fairness_bound}
\begin{aligned}
    &T \cdot g(\frac{1}{T}\sum_{t=1}^T A_t x_t) - \sum_{t=1}^T g(z_t)\\
    \leq & T L \| \frac{1}{T} \sum_{t=1}^T A_t x_t - \frac{1}{T}\sum_{t=1}^T z_t\|  + T \biggl[ g(\frac{1}{T}\sum_{t=1}^T z_t) - \frac{1}{T} \sum_{t=1}^T g(z_t) \biggr] \\
    \leq & L \| \sum_{t=1}^T A_t x_t - \sum_{t=1}^T z_t\| . 
\end{aligned}
\end{equation}
Since $\kappa_{t+1}$ is the minimizer of $e(\kappa)$, we have
\begin{equation}
    d_t + \frac{1}{\eta} \bigl( \nabla h(\kappa_{t+1}) - \nabla h(\kappa_{t}) \bigr) = 0 .
\end{equation}
     
In other words
\begin{equation}
    \Bigl\| \sum_{t=1}^T (- d_t) \Bigr\| = \Bigl\| \sum_{t=1}^T \bigl(A_t x_t - z_t \bigr) \Bigr\| =  \frac{\| \nabla h(\kappa_{T+1}) - \nabla h(\kappa_{1}) \| }{\eta}   .
\end{equation}
Since the reference function $h(\cdot)$ is $l$-strongly convex and $\beta_2$-smooth, we have
\begin{equation}
\begin{aligned}
    &\frac{\| \nabla h(\kappa_{T+1}) - \nabla h(\kappa_{1}) \| }{\eta} \leq \frac{\beta_2 \| \kappa_{T+1} -  \kappa_{1} \| }{\eta}  
    \leq  \frac{\beta_2}{\eta} \sqrt{\frac{2}{l} V_h(\kappa_1, \kappa_{T+1})} . 
\end{aligned}
\end{equation}
By substituting the inequality to Eqn~\eqref{eqn:fairness_bound}, we have
\begin{equation}
\begin{aligned}
    T \cdot g(\frac{1}{T}\sum_{t=1}^T A_t x_t) - \sum_{t=1}^T g(z_t)  \leq   \frac{\beta_2 L}{\eta} \sqrt{\frac{2}{l} V_h(\kappa_1, \kappa_{T+1})} . 
\end{aligned}
\end{equation}
By substituting Lemma \ref{lemma:queue_length_general_convex} into the inequality, we have
\begin{equation}
\begin{aligned}
     & T \cdot g(\frac{1}{T}\sum_{t=1}^T A_t x_t) - \sum_{t=1}^T g(z_t)   
     \leq {\beta_2 L} \sqrt{ T(  \frac{Z^2}{l^2} +  \frac{1}{\eta} \frac{2 L Z}{l} )-  \frac{2}{\eta^2 l} \kappa_1^\top \cdot (\sum_{t=1}^T A_t  x_t  - z_t)} .
\end{aligned}
\end{equation}
According to our assumption, we have $\Bigl| \kappa_1^\top \cdot (\sum_{t=1}^T A_t  x_t  - z_t) \Bigr| \leq T Z \cdot \| \kappa_1 \|$.
By substituting the above inequality into Eqn~\eqref{eqn:objective_difference_1}, we have
\begin{equation}
    \begin{aligned}
        & \sum_{t=1}^T \biggl[ f_t(x_t) + \lambda_1 d(x_t, x_{t-1}) + \frac{\lambda_2}{2} \| x_t - v_t \|^2    \biggr]  + T \cdot g(\frac{1}{T}\sum_{t=1}^T A_t x_t)\\
        \leq &   \sum_{t=1}^T \biggl[ f_t(x_t^*) + \lambda_1 d(x_t^*, x_{t-1}) + \frac{\lambda_2}{2} \| x_t^* - v_t \|^2 \biggr]  + T \cdot g(\frac{1}{T}\sum_{t=1}^{T} A_t x_t^*)    + \frac{\eta}{2l} Z^2 T R  \\
        & + TZ\|\kappa_1\| + L\delta  - \frac{m+\lambda_1  \beta_1 + \lambda_2}{2}\sum_{t=1}^T \| x_t^* - x_t\|^2 + {\beta_2 L} \sqrt{ T \Bigl( \frac{Z^2}{l^2} +  \frac{1}{\eta} \frac{2 L Z}{l}  + \frac{2}{\eta^2 l} Z \| \kappa_1 \| \Bigr) } .
    \end{aligned}
\end{equation}
By dividing both sides by $T$, we finish the proof.
 
\end{proof}

Now we are ready to prove Theorem~\ref{thm:dmd_algorithm_switch} as the following.
\begin{proof}

According to Lemma~\ref{lemma:intermediate_cost}, the switching cost in right hand side in Eqn~\eqref{eqn:cost_gap_1} is evaluated on the actual action sequence, while the ultimate goal is to bound it with the switching cost of offline optimal action sequence. In other words, the final step is to convert $ G({x}_{1:T}^*, {x}_{0:T-1})$ to $ G({x}_{1:T}^*, {x}_{0:T-1}^*)$. More specifically,

\begin{equation}
\begin{aligned}\label{eqn:switch_bound}
    & \lambda_1 d({x}_t, {x}_{t-1}) = \frac{\lambda_1  \beta_1}{2}\sum_{t=1}^T  \| x_t^* -  x_{t-1}\|^2 \\
    \leq & \frac{\lambda_1 \beta_1}{2}\sum_{t=1}^T  \biggl( \| x_t^* - x_{t-1}^*\| + \|x_{t-1}^*-  x_{t-1}\| \biggr)^2 \\
    \leq & \frac{\lambda_1  \beta_1}{2}(\frac{m+\lambda_2 + \lambda_1 \beta_1}{m+\lambda_2})\sum_{t=1}^T  \| x_t^* - x_{t-1}^*\|^2  + \frac{m+\lambda_1  \beta_1 + \lambda_2}{2}\sum_{t=1}^T  \|x_{t-1}^*-  x_{t-1}\|^2\\
    \leq & \frac{\lambda_1 \beta_1}{2}(\frac{m+\lambda_2 + \lambda_1 \beta_1}{m+\lambda_2})\sum_{t=1}^T  \| x_t^* - x_{t-1}^*\|^2   + \frac{m+\lambda_1 \beta_1 + \lambda_2}{2}\sum_{t=1}^T  \|x_{t}^*-  x_{t}\|^2 . 
\end{aligned}
\end{equation}
Substitute Eqn~\eqref{eqn:switch_bound} into Eqn~\eqref{eqn:cost_gap_1}, we have

\begin{equation}
\begin{aligned}
    &\frac{1}{T}\sum_{t=1}^T \biggl[ f_t({x}_t) + \lambda_1 d({x}_t, {x}_{t-1}) + \frac{\lambda_2}{2} \| {x}_t - v_t \|^2 \biggr]  +   g(\frac{1}{T}\sum_{t=1}^{T} A_t {x}_t)\\
    \leq & \frac{1}{T}\sum_{t=1}^T \biggl[ f_t({x}_t^*) + \lambda_1 (\frac{m+\lambda_2 + \lambda_1 \beta_1}{m+\lambda_2}) d({x}_t^*, {x}_{t-1}^*)  + \frac{\lambda_2}{2} \| {x}_t^* - v_t \|^2 \biggr] +  g(\frac{1}{T}\sum_{t=1}^{T} A_t {x}_t^*) + 0 + \Delta \\
    \leq & \frac{1}{T}\sum_{t=1}^T \biggl[ (1 + \frac{\lambda_2}{m})f_t({x}_t^*) + \lambda_1 (\frac{m+\lambda_2 + \lambda_1 \beta_1}{m+\lambda_2}) d({x}_t^*, {x}_{t-1}^*) \biggr]  + g(\frac{1}{T}\sum_{t=1}^{T} A_t {x}_t^*) + \Delta
\end{aligned}
\end{equation}
where $\Delta = \frac{\eta}{2l} Z^2 R + \beta_2  L \sqrt{\frac{1}{T}(\frac{Z^2}{l^2} + 2\frac{L Z}{\eta l} + \frac{2}{\eta^2 l} Z \| \kappa_1\|)}+ Z \|\kappa_1 \| + \frac{L\delta}{T} $. 
 
Then the overall cost of \ouralg is bounded by 
\begin{equation}
\begin{aligned}
    & \frac{1}{T}\sum_{t=1}^T \biggl[ f_t({x}_t) +  d({x}_t, {x}_{t-1}) \biggr] +   g(\frac{1}{T}\sum_{t=1}^{T} A_t {x}_t)  \\
    \leq & \frac{1}{T\lambda_1}\sum_{t=1}^T \biggl[ f_t({x}_t) + \lambda_1 d({x}_t, {x}_{t-1}) \biggr] +   \frac{1}{\lambda_1}g(\frac{1}{T}\sum_{t=1}^{T} A_t {x}_t)  + \frac{1}{\lambda_1}\Delta\\
    \leq &  \frac{1}{T}\sum_{t=1}^T \biggl[ (\frac{m + \lambda_2}{m\lambda_1})f_t({x}_t^*) +  (\frac{m+\lambda_2 + \lambda_1 \beta_1}{m+\lambda_2}) d({x}_t^*, {x}_{t-1}^*) \biggr] + \frac{1}{\lambda_1} g(\frac{1}{T}\sum_{t=1}^{T} A_t {x}_t^*) + \frac{1}{\lambda_1}\Delta \\
    \leq & C \Biggl[  \frac{1}{T}\sum_{t=1}^T \biggl[f_t({x}_t^*) +  d({x}_t^*, {x}_{t-1}^*) \biggr] +  g(\frac{1}{T}\sum_{t=1}^{T} A_t {x}_t^*) \Biggr] + \frac{1}{\lambda_1}\Delta  . 
\end{aligned}
\end{equation}
where $C = \max\{ \frac{m + \lambda_2}{m\lambda_1}, \frac{m+\lambda_2 + \lambda_1 \beta_1}{m+\lambda_2} \}$ is the competitive ratio. 
\end{proof}